\documentclass[runningheads]{llncs}

\usepackage[year=2026]{eccv}

\usepackage{eccvabbrv}
\usepackage{graphicx}
\usepackage{booktabs}
\usepackage{multirow}
\usepackage{tabularx}
\usepackage{pifont} 
\usepackage{array}
\usepackage{appendix}
\usepackage{enumitem}
\usepackage{algorithm}
\usepackage{algpseudocode}
\newcolumntype{Y}{>{\centering\arraybackslash}X}

\usepackage[table]{xcolor}
\usepackage[accsupp]{axessibility}  

\usepackage{hyperref}

\usepackage{orcidlink}
\usepackage[utf8]{inputenc}

\begin{document}

\title{Cross-Resolution Distribution Matching for Diffusion Distillation}

\titlerunning{Cross-Resolution Distribution Matching for Diffusion Distillation}

\author{
Feiyang Chen\inst{1,3}\textsuperscript{*,\ddag} \and
Hongpeng Pan\inst{1}\textsuperscript{*} \and
Haonan Xu\inst{1,2}\textsuperscript{\ddag} \and
Xinyu Duan\inst{1} \and
 Zhefeng Wang\inst{1}\textsuperscript{\dag} \and 
 Yang Yang\inst{2}\textsuperscript{\dag}
}

\authorrunning{F.~Chen et al.}

\institute{Huawei, Shanghai \& Hangzhou, China \and Nanjing University of Science and Technology, Nanjing, China \and HiThink Research, Hangzhou, China }
\maketitle
\insert\footins{\footnotesize \textsuperscript{*}Equal contribution.}
\insert\footins{\footnotesize \textsuperscript{\dag}Corresponding authors. Email: yyang@njust.edu.cn, wangzhefeng@huawei.com}
\insert\footins{\footnotesize \textsuperscript{\ddag}Work done while Feiyang Chen and Haonan Xu were with Huawei.}
\begin{abstract}
Diffusion distillation is central to accelerating image and video generation, yet existing methods are fundamentally limited by the denoising process, where step reduction has largely saturated. Partial-timestep, low-resolution generation can further accelerate inference, but it suffers from noticeable quality degradation due to cross-resolution distribution gaps. We propose Cross-\textbf{R}esolution Distribution \textbf{M}atching \textbf{D}istillation (RMD), a novel distillation framework that bridges cross-resolution distribution gaps for high-fidelity, few-step multi-resolution cascaded inference. 
Specifically, RMD divides the timestep intervals for each resolution using logarithmic signal-to-noise ratio (logSNR) curves, and introduces logSNR-based mapping to compensate for resolution-induced shifts. Distribution matching is conducted along resolution trajectories to reduce the gap between low-resolution generator distributions and the teacher’s high-resolution distribution. In addition, a predicted-noise re-injection mechanism is incorporated during upsampling to stabilize training and improve synthesis quality. Quantitative and qualitative results show that RMD preserves high-fidelity generation while accelerating inference across various backbones. Notably, RMD achieves up to $33.4\times$ speedup on SDXL and $25.6\times$ on Wan2.1-14B, while preserving high visual fidelity.


\end{abstract}

\section{Introduction}
\label{sec:intro}
In recent years, diffusion models \cite{Ho2020DDPM, Lipman2023FM} have demonstrated impressive performance in high-fidelity visual generation \cite{Rombach2022LDM,Wan2025Video,Hunyuan2025Video}. Despite their success, diffusion models require hundreds of iterative denoising steps, each involving a full forward pass through a large-scale neural network. Moreover, in modern Diffusion Transformer \cite{Peebles2023DiT} (DiT) architectures, the computational cost of each forward pass is further amplified at high resolutions \cite{Yuan2024DiTFastAttn}, as the number of tokens scales with spatial dimensions, leading to a quadratic increase in attention complexity. As a result, visual synthesis incurs substantial computational overhead and latency, limiting the feasibility of real-time or resource-constrained applications.

Up to now, many efforts have been made in pursuing efficient sampling techniques \cite{Shen2025Survey}. Among them, training-free methods \cite{Song2021DDIM,Lu2025DPM-Solver++,Liu2025TE-Cache} accelerate sampling but still need dozens of steps. In contrast, step distillation methods, such as trajectory-based distillation \cite{Salimans2022PD,Song2023CM,Kim2024CTM} and distribution-matching distillation \cite{Yin2024DMD,Yin2024DMD2,Zhou2024SiD}, can significantly improve sampling efficiency, reducing the process to only a few steps (e.g., to 4–8 steps) while maintaining high-fidelity visual generation. However, empirical evidence \cite{Luo2025TDM} indicates that excessively aggressive reduction of generation steps (e.g., to 1–3 steps) can result in a catastrophic decrease in performance. Consequently, relying solely on step reduction via distillation imposes inherent limitations on improving sampling efficiency.

To address the efficiency bottleneck in step distillation, we instead focus on lowering the  resolution at selected timesteps during the sampling process. Prior work has shown that the role of each denoising step depends on the noise level: early steps under high-noise conditions primarily reconstruct the global structural layout, whereas later steps under low-noise conditions refine fine-grained visual details~\cite{Wang2023Painter, SDEdit, Zhang2025FlashVideo}. However, performing high-resolution reconstruction during the high-noise stage leads to redundant computation of fine-grained details that are not yet required. Therefore, to further improve the efficiency of step distillation, we adopt a multi-resolution cascaded generation strategy: global structures are first recovered via coarse denoising at a lower resolution, and high-resolution processing is deferred to later stages to refine fine-grained details.

\begin{figure}[t]
    \centering
    \includegraphics[width=1\textwidth]{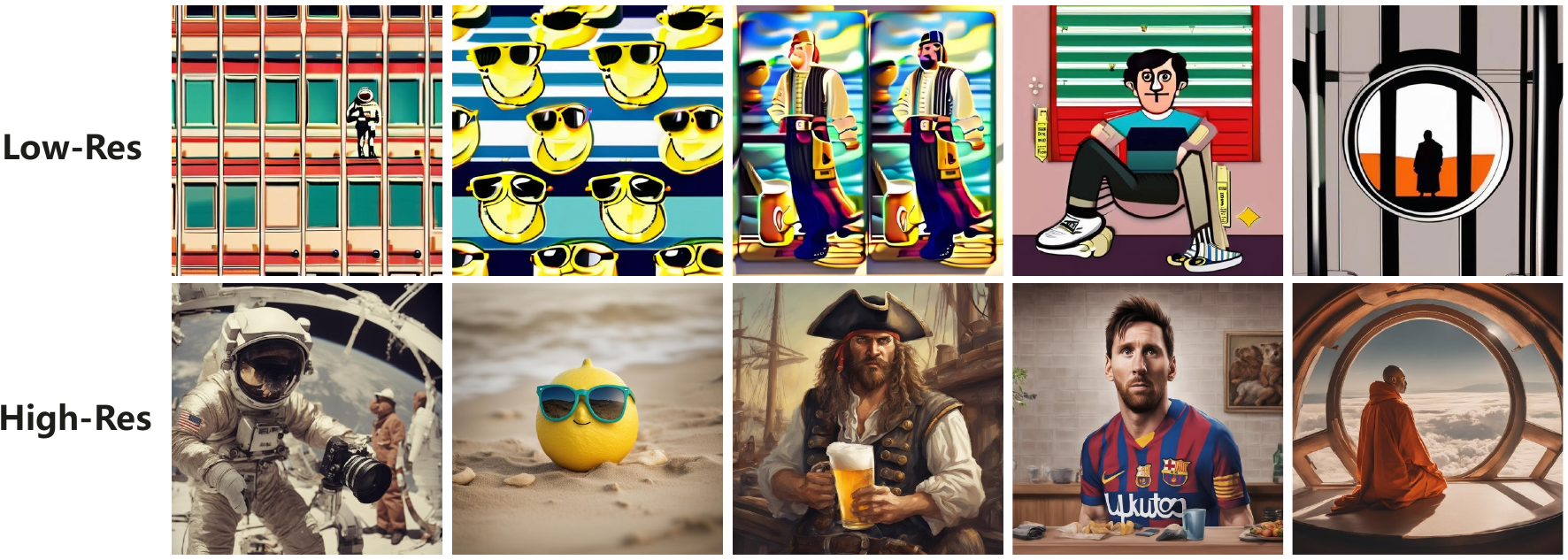}
    \caption{
        Under the same text prompt and random seed, the SDXL model produces distinct images at 512×512 and 1024×1024 resolutions, revealing a clear resolution-dependent distribution shift. The corresponding prompts are provided in Appendix D.
    }
    \label{fig:intro}
\end{figure}

However, as illustrated in Fig.~\ref{fig:intro}, the same model yields distinct data distributions across resolutions; thus, directly reducing the generation resolution at selected timesteps introduces distributional inconsistencies and degrades synthesis quality~\cite{Ma2025NAMI,Tian2025BottleneckSampling}. This issue stems from the training paradigm of existing state-of-the-art diffusion models~\cite{Hunyuan2025Video, Wan2025Video, Zhang2025Waver}, which adopt a multi-stage curriculum schedule: models are first trained on large-scale, low-resolution data of varying quality, and subsequently fine-tuned on carefully curated high-resolution data. Consequently, under a coarse-to-fine multi-resolution cascaded generation scheme, the global structure is effectively sampled from a low-resolution distribution with heterogeneous quality, rather than from the inherently higher-fidelity high-resolution distribution, resulting in a fundamental mismatch.

In this paper, we introduce RMD, a novel distillation framework (see Fig.~\ref{fig:intro_2}) for few-step multi-resolution cascaded generation that explicitly bridges cross-resolution distribution inconsistencies. Concretely, RMD segments the denoising trajectory according to logSNR curves and progressively increases the resolution as noise decreases, enabling coarse-grained semantic construction at early stages and fine-grained detail refinement at later stages. Within each segment, noisy samples are upsampled and aligned at the final high resolution for distribution matching distillation, ensuring global distribution consistency across resolutions. In addition, a predicted-noise re-injection mechanism is employed during upsampling to provide prior trajectory guidance, stabilizing training and enhancing synthesis quality. As a result, RMD achieves few-step generation with progressive resolution scaling, breaking through the sampling efficiency bottleneck of conventional step-distillation methods, while preserving high-fidelity visual synthesis. Extensive experiments on image synthesis and its natural extension to video synthesis validate the effectiveness of the proposed RMD method, consistently achieving improved sampling speed while preserving high visual fidelity.

\section{Related Work}
\subsection{Step Distillation}

\noindent \textbf{Trajectory-based Distillation} trains a few-step student model by aligning its denoising trajectory with that of a multi-step teacher model \cite{Meng2023GuidedDistill,Song2023CM,Liu2024InstaFlow}. Progressive distillation \cite{Salimans2022PD} iteratively trains a sequence of student models, each halving the number of sampling steps of its predecessor by matching two-step trajectories in a single step. Instaflow \cite{Liu2024InstaFlow} straightens ODE trajectories to reduce discretization error and enable generation with fewer steps. Consistency models \cite{Song2023CM} and their subsequent extensions \cite{Wang2024PCM,Song2024ITC,Kim2024CTM,Wang2024PCM} train student models so that their outputs remain self-consistent at any timestep along the ODE trajectory. However, these methods suffer from difficulties in instance-level trajectory matching and numerical errors when solving the teacher’s probability flow ODE \cite{Luo2025TDM}.

\noindent \textbf{Distribution Matching Distillation (DMD) \cite{Yin2024DMD}} trains the student by aligning its output distribution with the multi-step teacher, avoiding the difficulty of instance-level trajectory matching. To date, many follow-up works \cite{Xu2025fDMD,Bandyopadhyay2025SD35Flash,Lu2025ADMD,Zhou2024SiD,Jiang2025RLDMD} have proposed improvements. For example, DMD2 \cite{Yin2024DMD2} introduces a discriminator to align the student model’s distribution with a specified target distribution in a GAN-like manner \cite{goodfellow2020generative}. TDM \cite{Luo2025TDM} enforces distribution-level alignment between the student’s trajectory and that of the teacher. DP-DMD \cite{wu2026diversity} exempts the first step from DMD, preserving sample diversity while enabling efficient, high-quality generation. Our work also starts with DMD, where we incorporate a coarse-to-fine generation paradigm into end-to-end distillation, going beyond the efficiency limits of step-only compression.

\subsection{Multi-Resolution Cascaded Methods}
Multi-resolution cascaded methods accelerate diffusion-based generation by decomposing the denoising process into a multi-resolution hierarchy, proceeding in a coarse-to-fine manner~
\cite{Zhang2025DTC, Zhang2025FlashVideo, Ma2025NAMI, Teng2024Relay}. For instance, PixelFlow \cite{Chen2025PixelFlow} is trained from scratch to model cascaded flows in the pixel space, allowing images to be progressively synthesized from low to high resolution. Bottleneck Sampling \cite{Tian2025BottleneckSampling} employs a high-low-high denoising workflow to maintain both visual fidelity and sampling efficiency. Moreover, cascade strategy has been incorporated into large-scale state-of-the-art diffusion models \cite{Zhang2025Waver,Cai2025LongCat}, employing cascade refiners to facilitate efficient high-resolution visual synthesis. In this work, we unify multi-resolution cascading into a step distillation framework and address the distribution inconsistency between low- and high-resolution cascaded denoising \cite{Tian2025BottleneckSampling}.

\begin{figure}[t]
    \centering
    \begin{minipage}[t]{0.54\textwidth}
        \centering
        \includegraphics[width=\textwidth]{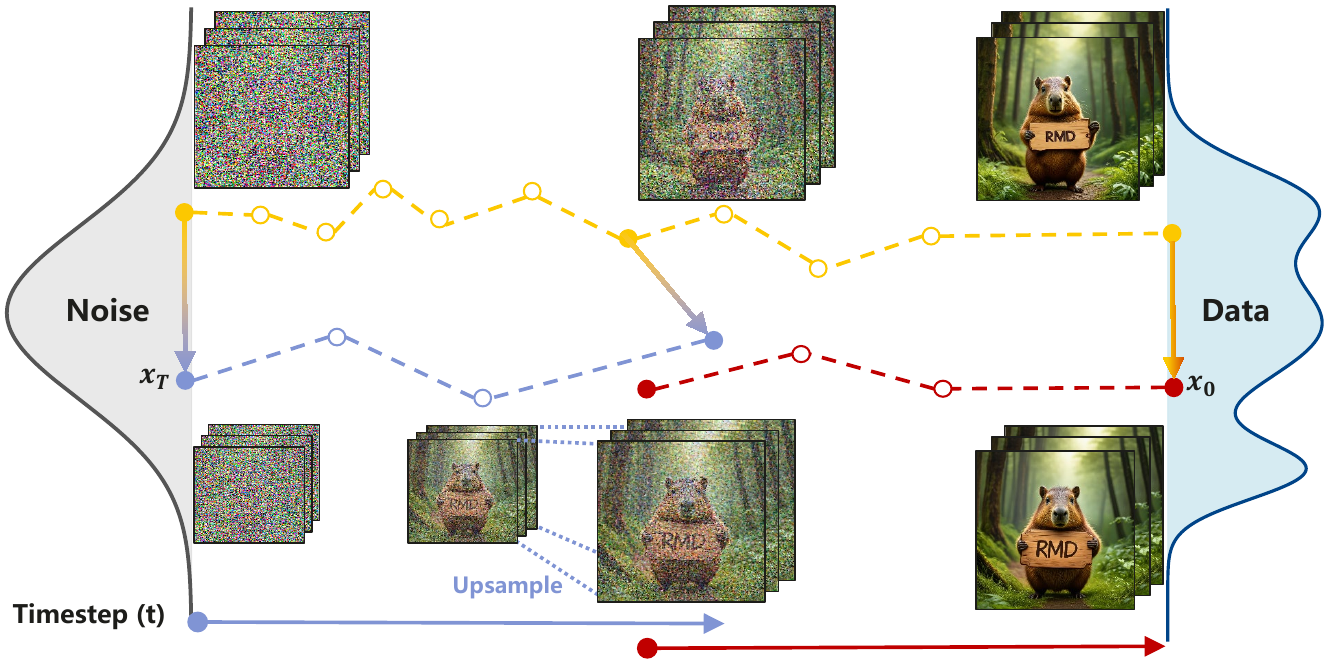}

        \caption{Overview of the RMD Framework, which compresses the denoising trajectory of a pretrained diffusion model into a multi-resolution, few-step cascaded denoising process.}
        \label{fig:intro_2}
    \end{minipage}
    \hfill
    \begin{minipage}[t]{0.44\textwidth}
        \centering
        \includegraphics[width=\textwidth]{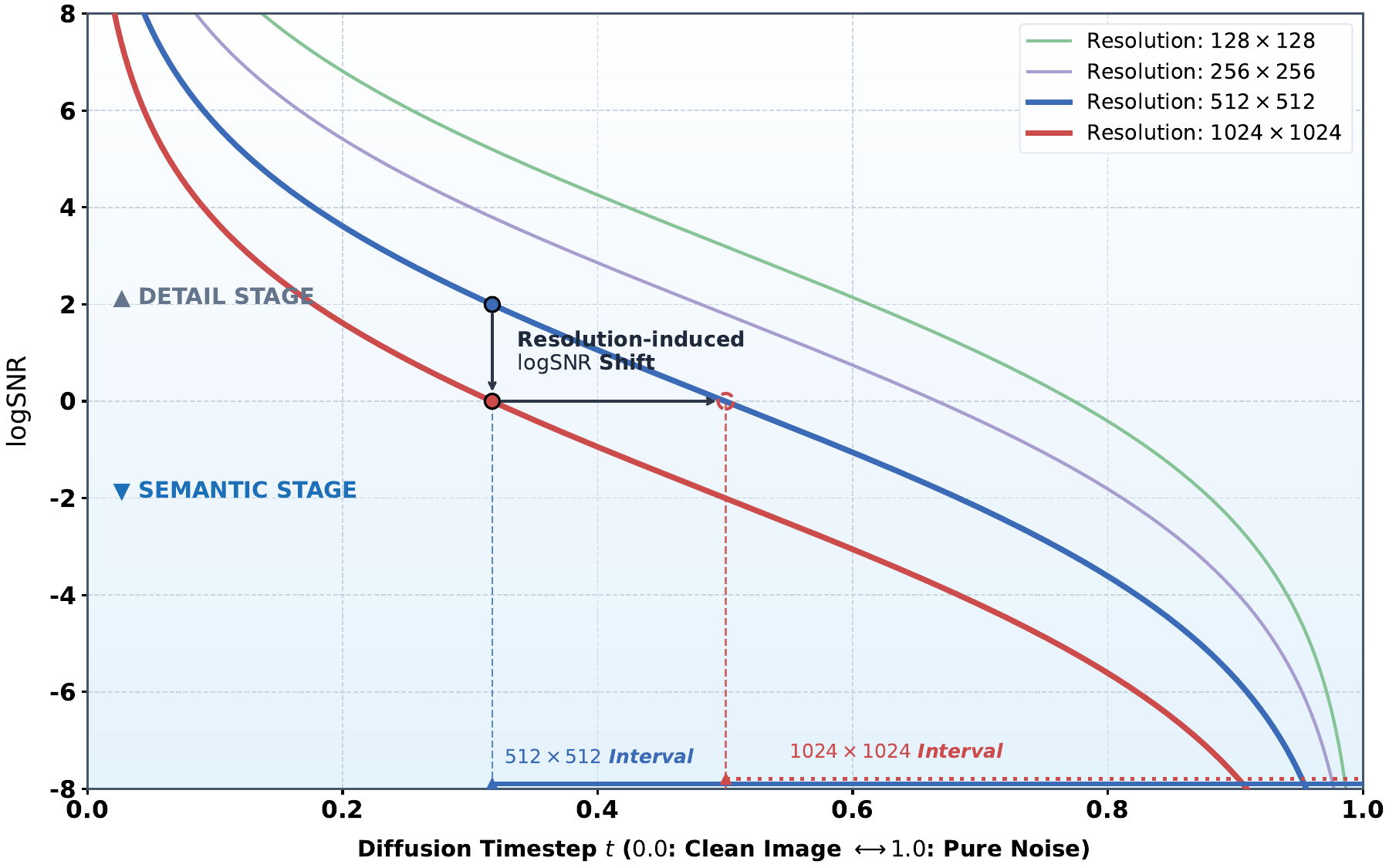} 
  
        \caption{Illustration of Cross-Resolution Timestep Interval Alignment. logSNR curves at different resolutions show resolution-dependent variations in noising dynamics.}
        \label{fig:Meth_1}
    \end{minipage}

\end{figure}

\section{Methodology}
Cross-Resolution Distribution Matching Distillation (RMD) is introduced as a novel distillation framework that aligns the student’s low-resolution distribution with the teacher’s high-resolution distribution at the distribution level. We first partition the diffusion timestep into resolution-specific intervals based on their log signal-to-noise ratio (logSNR) profiles. A cross-resolution distribution matching objective is then formulated to learn distributional matching across resolutions. Finally, an optimized upsampling strategy is adopted to reduce the difficulty of cross-resolution alignment and improve training effectiveness.
\begin{figure}[!t]
  \centering
  \includegraphics[width=0.8\linewidth]{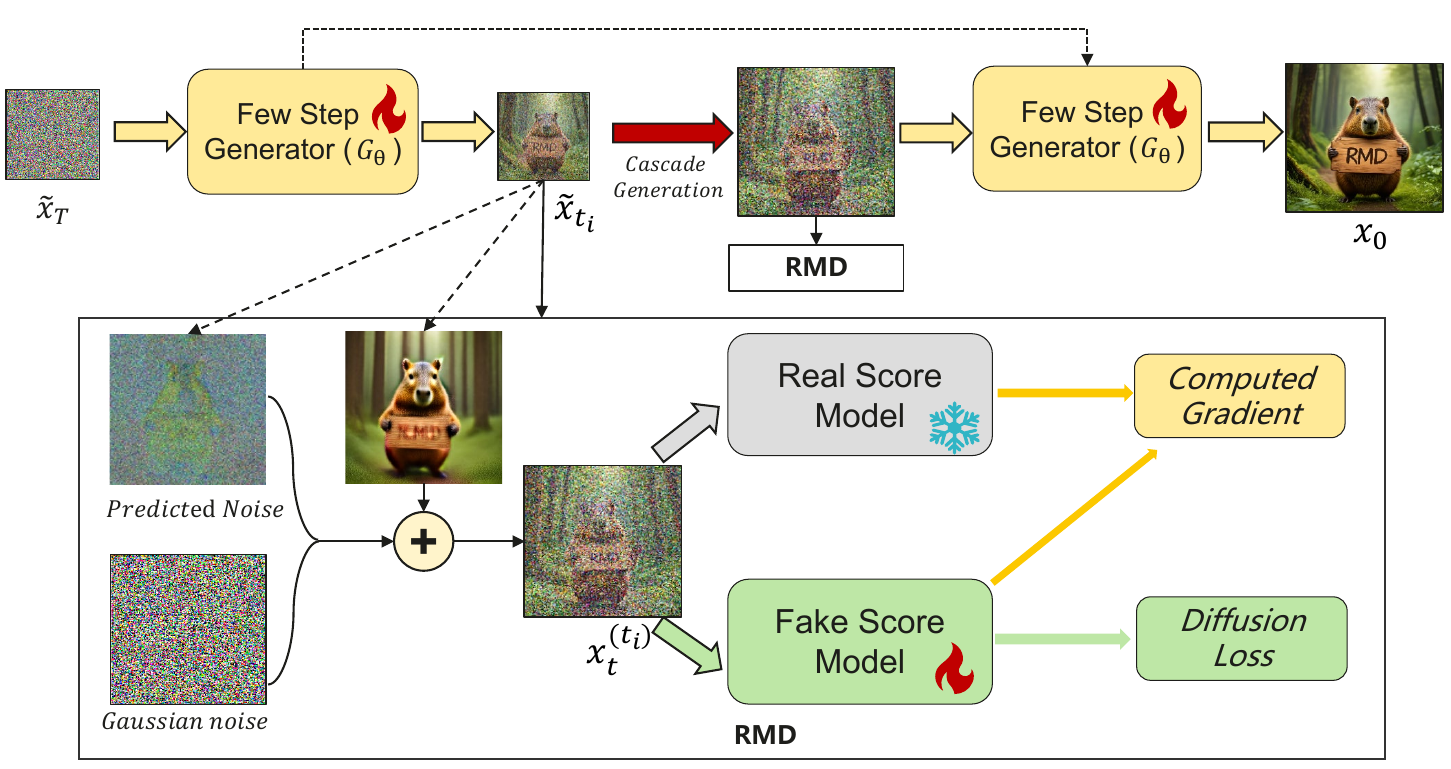} 
  \caption{The overall pipeline for distilling a pretrained diffusion model into a cascaded generator $G_\theta$ that performs generation across two resolution distribution spaces.}
  \label{fig:method_framework_2}

\end{figure}
\subsection{Resolution Trajectory Division}

Resolution scaling intrinsically modifies the effective noise schedule. For identical independent noise magnitude, higher resolutions suffer relatively weaker pixel corruption~\cite{Hoogeboom2023SimpleDiffusion, Chen2023NoiseSchedule, Teng2024Relay}. This effect can be characterized via the logSNR. As shown in Fig.~\ref{fig:Meth_1}, the growth rate of logSNR differs across resolutions: in the low-logSNR regime, low-resolution trajectories denoise more rapidly, whereas in the high-logSNR regime, higher resolutions achieve a higher denoising rate. This observation motivates partitioning the diffusion process according to the logSNR trajectory and assigning resolution-specific timestep intervals. Note that Fig.~\ref{fig:Meth_1} plots the \emph{compensated} logSNR schedules after applying Eq.~\eqref{con:logsnr_shift}, where low-resolution curves sit above high-resolution curves, indicating higher compensated logSNR (less effective noise) at each timestep.

To ensure distributional alignment with the teacher model, we partition the teacher’s logSNR curve at predefined thresholds $\{\text{logSNR}_1, \dots, \text{logSNR}_{K-1}\}$, yielding $K$ non-overlapping diffusion segments. Each breakpoint is converted to a diffusion boundary via the logSNR–timestep mapping. Under the Rectified Flow parameterization~\cite{Liu2023FSF}, the conversion is
\begin{equation}
\label{con:logsnr}
\sigma_{T_i} = \frac{1}{1+\exp\left( \frac{\text{logSNR}_i}{2} \right)}, 
\end{equation}
where $\sigma_{T_i}$ denotes the noise schedule value at the boundary timestep $T_i$, with $\sigma_{T_0}=1$ and $\sigma_{T_K}=0$~\cite{Liu2023FSF}, corresponding to the diffusion timestep partition defined by the logSNR curve. Each segment $i\in\{1,2,\dots, K\}$ corresponds to a timestep interval $\left [ T_{i-1}, T_{i} \right ]$, and is assigned a predefined resolution $r_i$, satisfying $r_1 < r_2 < \cdots < r_{K}$, where $r_{K}$ equals the teacher model’s generation resolution. 

To maintain consistency with the teacher's timestep intervals during RMD, we compute the timestep shifts induced by resolution variations. Following the resolution-dependent scaling in SD3~\cite{Esser2024SD3}, the adjusted logSNR at resolution $r_i$ is formulated as:
\begin{equation}
\label{con:logsnr_shift}
\text{logSNR}^{(r_i)}_i
=
\text{logSNR}_i
- 2 \log\!\left( \frac{r_i}{r_K} \right),
\end{equation}
where the second term compensates for the resolution-induced SNR shift. By substituting the adjusted logSNR into Eq.~\eqref{con:logsnr}, we derive the student's timestep interval $[T^{(r_{i})}_{i-1}, T^{(r_{i})}_{i}]$ that corresponds to the teacher’s segment $[T_{i-1}, T_{i}]$. As illustrated in Fig.~\ref{fig:Meth_1}, we exemplify this for a two-stage ($K=2$) configuration: by mapping the high-resolution interval onto the low-resolution temporal axis through logSNR invariance, we can bridge the gap between varying scales. This temporal synchronization ensures that distillation occurs over identical denoising states, regardless of the spatial resolution.

\subsection{Cross-Resolution Distribution Matching}

The objective of RMD is to compress a pretrained diffusion model into a few-step, multi-resolution cascaded generator $G_\theta$. For each resolution interval, the generator is trained to match the teacher distribution under aligned logSNRs. Formally, we minimize the joint KL divergence~\cite{Zhou2024SiD, Yin2024DMD, Yin2024DMD2}:

\begin{equation}
\label{con:KL_divergence}
\min_\theta
\mathrm{KL}\Big(
p_{\theta}(\tilde{x}_{t_i}) \, p_{\theta}(x_{0} \mid \tilde{x}_{t_i}) \;\|\; p_{\varphi}(x_t) \, p_{\varphi }(x_{0} \mid x_t)
\Big),
\end{equation}
where $p_{\varphi}$ and $p_{\theta}$ denote the pretrained teacher and generator, respectively. The index $t_i := (r_i, t)$ represents the resolution-aware timestep corresponding to teacher timestep $t$ with resolution $r_i$ under matched logSNR, as in Eq.~\eqref{con:logsnr_shift}. $\tilde{x}_{t_i}$ denotes the generator state aligned with the teacher distribution $x_t$. The conditionals $p_{\varphi}(x_{0} \mid x_t)$ and $p_{\theta}(x_{0} \mid \tilde{x}_{t_i})$ correspond to teacher ODE denoising and generator cross-resolution denoising.

Optimizing the above objective introduces two challenges. 
First, point-wise distribution matching along the generation trajectory is insufficient for effective distillation~\cite{Luo2025TDM}. 
Second, as $x_{0}$ and $\tilde{x}_{t_i}$ reside at different resolutions, directly denoising $\tilde{x}_{t_i}$ to recover $x_{0}$ is ill-posed. Moreover, $\tilde{x}_{t_i}$ contains stochastic diffusion noise, and naive upsampling may distort low-resolution structural priors.

To overcome these issues, we project the generator state $\tilde{x}_{t_i}$ into the teacher space via a differentiable upsampling transformation~\cite{Zhang2025DTC, Tian2025BottleneckSampling}, and perform distribution level matching along the inference trajectory. Specifically, the teacher scheduler determines $N$ inference steps $\{t_j\}_{j=0}^{N-1}$, uniformly defined as $t_j = T_0-\frac{T_0}{N}j$. 
These steps are mapped to low-resolution timestep as $t_{i,j} := (r_i, t_j)$. 
Multi-resolution cascaded inference under $G_{\theta}$ produces intermediate states $\tilde{x}_{t_{i,j}}$, which are subsequently transformed to the high-resolution space via the following upsampling transformation:

\begin{equation}
\begin{gathered}
\tilde{x}^{(t_{i,j})}_0
= \tilde{x}_{t_{i,j}} 
- \sigma_{t_{i,j}}\, \epsilon_{\theta}(\tilde{x}_{t_{i,j}}, t_{i,j}), \\
x_{t_j}^{(t_{i,j})} 
= (1-\sigma_{t_{j}}) \, U_{r_{K}}(\tilde{x}^{(t_{i,j})}_0) 
+ \sigma_{t_{j}} \, \epsilon .
\end{gathered}
\end{equation}

Here, $\epsilon_{\theta}$ denotes the noise prediction of $G_{\theta}$, and $U_{r_{K}}(\cdot)$ denotes an interpolation upsampling operator to resolution $r_K$. The time-dependent standard deviation $\sigma_t$ controls the noise scale, and $\epsilon \sim \mathcal{N}(0,I)$ represents the injected stochastic noise. $\tilde{x}^{(t_{i,j})}_0$ denotes the clean latent from the low-resolution distribution at timestep $t_{i,j}$. For simplicity, we adopt a rectified flow formulation~\cite{Liu2023FSF} for noise injection, though other strategies~\cite{Ho2020DDPM, Lipman2023FM} are readily applicable. 

Substituting the induced distribution of $x_{t_j}^{(t_{i,j})}$ into the distillation objective, we minimize the marginal reverse KL divergence under score distillation~\cite{Diff-Instruct, Yin2024DMD}:
\begin{equation}
L(\theta) = \sum_{j=1}^{N} 
\mathrm{KL}\Big(
p_{\theta}(x^{(t_{i,j})}_{t_j}) \;\|\; p_{\varphi}(x_{t_j})\Big).
\label{eq:kl_loss}
\end{equation}

However, multi-step alignment across resolutions is computationally demanding. Furthermore, because different resolutions require varying numbers of sampling steps, the resulting supervision becomes imbalanced, leading to slow convergence. Consequently, we perform distribution matching exclusively along resolution-wise stages. For each resolution stage $r_i$, we uniformly sample timesteps:  
\[
t_i \sim \mathcal{U}([T_{i-1}^{(r_i)}, T_i^{(r_i)}]).
\]

We denote the corresponding expectation simply as $\mathbb{E}_{t_i}$ for brevity, and perform cross-resolution distribution alignment at the sampled steps. 
The overall objective is formulated as: 
\begin{equation}
L(\theta) = \sum_{i=1}^{K} \lambda_{r_i} \, \mathbb{E}_{t_i} \Big[ \mathrm{KL}\big( p_\theta(x_t^{(t_i)}) \;\|\; p_\varphi(x_t) \big) \Big].
\end{equation}

Here $p_\theta(x^{(t_i)}_t)$ denotes the marginal diffusion distribution at timestep $t_i$ within resolution stage $r_i$, projected to the high-resolution space at timestep $t$. The coefficient $\lambda_{r_i}$ is a resolution-aware weight that balances the contribution of each resolution interval in the optimization objective. In practice, the expectation is approximated via Monte Carlo sampling during training.

Using score approximation, the gradient can be written as:
\begin{equation}
\nabla_{\theta} L(\theta) \approx 
\sum_{i=1}^{K} 
\lambda_{r_i}
\mathbb{E}_{t_i}
\Big[
\big(
s_{\theta}(x^{(t_i)}_t, t)
-
s_{\varphi}(x_{t}, t)
\big)
\frac{\partial x^{(t_i)}_t}{\partial\theta}
\Big].
\end{equation}

Here, $x^{(t_i)}_t$ denotes the upsampled distribution obtained from the generator outputs via the upsampling transformation, which remains differentiable with respect to the model parameters $\theta$. 

The generator score $s_{\theta}$ is intractable. Following DMD~\cite{Yin2024DMD} and TDM~\cite{Luo2025TDM} score approximation, we introduce a fake diffusion model $s_{\phi}$ to estimate it, termed the fake score, while $s_{\varphi}$ is treated as the true score. With the introduced fake diffusion model $s_{\phi}$, the final gradient update formulation for RMD is:
\begin{equation}
\label{eq:score-distillation-loss}
L(\theta) =
\sum_{i=1}^{K} \, 
\lambda_{r_i}
\mathbb{E}_{t_i}
\Big\|
x_t^{(t_i)} 
- \operatorname{sg} \big(
    x_t^{(t_i)} 
    + s_{\phi}(x_t^{(t_i)}, t)
    - s_{\varphi}(x_t, t)
\big)
\Big\|_2^2 \,.
\end{equation}

Here, $\mathrm{sg}(\cdot)$ denotes the stop-gradient operator, which can be interpreted as the upsampled generator distribution after being refined by one step of gradient descent, while blocking gradient back-propagation. A concise overview of the cross-resolution matching distillation training procedure is shown in Fig.~\ref{fig:method_framework_2}.

The fake diffusion model is initialized from a pretrained DM and optimized with a standard denoising objective to track score variations along generated trajectories:
\begin{equation}
L(\phi ) = 
\sum_{i=1}^{K}
\lambda_{\mathrm{snr}}
\mathbb{E}_{t_i}
\Big\|
f_{\phi}(x^{(t_i)}_t, t)
-
U_{r_{K}}(\tilde{x}^{(t_i)}_0)
\Big\|^2_2\;,
\label{eq:fake}
\end{equation}
where $U_{r_{K}}(\tilde{x}^{(t_i)}_0)$ is the clean target. $f_{\phi}$ denotes the denoiser network parameterized to predict the clean data. $\lambda_{\mathrm{snr}}$ is the resolution-aware denoising weight, defined as the SNR at timestep $t_i$ and resolution $r_i$.

\subsection{Upsampling Transformation Optimization}

In the upsampling transformation, the low-resolution distribution is upsampled and perturbed with Gaussian noise, enabling approximate alignment with the high-resolution distribution during training. However, such stochastic perturbation requires re-solving SDE trajectories across resolutions and does not inherit the teacher’s ODE trajectory, which increases the difficulty of RMD. In contrast, injecting purely predicted noise to mimic the teacher ODE trajectory leads to severe degradation when the resolution gap is large.

To address this issue, we introduce a noise re-injection strategy for RMD, where Gaussian noise is replaced by a resolution-aware combination of predicted and Gaussian noise. The revised formulation is:

\begin{equation}
\label{eq:noise_inject}
\begin{aligned}
\epsilon_{t_i}
&= \alpha \, U_{r_K}\!\left(
\epsilon_{\theta}(\tilde{x}_{t_i}, t_i)
\right)
+ \beta \, \epsilon \\
x^{(t_i)}_t 
&= (1-\sigma_{t})\, U_{r_K}\!(\tilde{x}^{(t_i)}_0)
+ \sigma_{t}\, \epsilon_{t_i}
\end{aligned}
\end{equation}

Here, $\epsilon_{t_i}$ comprises predicted noise and Gaussian noise. The predicted component is obtained by upsampling the generator’s noise prediction $\epsilon_{\theta}$, while the stochastic term is a Gaussian noise $\epsilon \sim \mathcal{N}(0, I)$. The two components are balanced by $\alpha$ and $\beta$ with $\alpha^2+\beta^2=1$. As the resolution gap enlarges, stronger stochasticity is required to bridge the distribution mismatch. Accordingly, $\alpha$ is reduced, assigning greater weight to the Gaussian noise.

\subsection{Training and Inference}

\noindent \textbf{Warm-up Training with Low logSNR}
The distribution induced by low-resolution generative trajectories largely determines the overall layout of the high-resolution distribution during distillation~\cite{SDEdit, Zhang2025FlashVideo, Zhang2025DTC}. In RMD, significant misalignment between the generated low-resolution distribution and the target high-resolution distribution can hinder the convergence of cascaded generative trajectories. To enable efficient optimization, the generative process is divided into semantic and detail generation intervals based on the logSNR partitioning scheme from Wan2.2. Specifically, a threshold $\text{logSNR}_{K/2}$ is selected to separate these intervals, and the low-logSNR (semantic) interval is first distilled in a warm-up phase to provide a stable initialization. The full trajectory is then trained end-to-end, accelerating the overall distillation process. The pseudocode for RMD training is provided in Algorithm 1 in Appendix A.

\noindent \textbf{Multi-Resolution Cascaded Inference.}
During inference, a multi-resolution cascaded scheme is adopted to synthesize the target-resolution distribution. The procedure dynamically determines whether upsampling is required at each subsequent timestep and, if so, re-injects noise to maintain consistency with the diffusion trajectory. Denoising begins from Gaussian noise at the lowest resolution $r_1$ and progressively increases the resolution of the generated distribution until the target resolution is reached. Within each stage, if the resolution remains unchanged at the next timestep, standard conditional diffusion denoising is performed. When the resolution changes, the current-stage distribution is first projected to the target stage, and then upsampling and noise are re-injected to match the higher-resolution distribution corresponding to the next timestep. This multi-resolution cascaded inference ensures seamless resolution transitions across stages while preserving the temporal consistency of the generation process. The complete pseudocode is provided in Algorithm 2 in Appendix A.

\section{Experiments}
\subsection{Experiments Setup}
\textbf{Baseline}. To assess the generalizability of RMD, we conduct extensive experiments across both image and video generation tasks. For text-to-image synthesis, we evaluate our method on various representative architectures, including the UNet-based Stable Diffusion XL (SDXL)~\cite{PodellELBDMPR2024SDXL} as well as the Transformer-based PixArt-$\alpha$~\cite{Chen2024Pixart} and Stable Diffusion 3.5 medium (SD3.5)~\cite{Esser2024SD3}. Specifically, we evaluate our method against two categories of publicly available distillation frameworks: (1) adversarial distillation, including SDXL-Turbo~\cite{Sauer2024ADD} and SDXL-Lighting~\cite{Shan2024PADD} (for SDXL), YOSO~\cite{LuoCQHT25YOSO} (for PixArt-$\alpha$), and SD3.5-Turbo~\cite{Bandyopadhyay2025SD35Flash} (for SD3.5); and (2) distribution matching methods, such as DMD2~\cite{Yin2024DMD2} and TDM~\cite{Luo2025TDM}. Furthermore, to demonstrate scalability to large-scale video models, we extend our evaluation to Wan2.1-T2V-14B~\cite{Wan2025Video}, where we conduct a comparison with distribution matching approaches.

\noindent \textbf{Metrics}. For image generation, following TDM~\cite{Luo2025TDM}, we use Human Preference Score v2.1 (HPS)~\cite{Wu2023HPS} for human-centric alignment, Aesthetic Score (AeS)~\cite{Schuhmann2022Aes} for perceptual beauty, and CLIP Score for text-visual semantic consistency. These metrics are cross-validated on images generated from the HPS v2.1 prompt set. For video generation, following Bottleneck Sampling~\cite{Tian2025BottleneckSampling}, we use VBench~\cite{Huang2024Vbench} to evaluate comprehensive video quality (e.g., temporal consistency and motion smoothness) and T2V-Compbench~\cite{Sun2025Compbench} to assess compositional capabilities. For both benchmarks, we use their provided original prompts for video generation. Furthermore, an extensive user study is conducted, with details provided in Appendix C.

\noindent \textbf{Implementation Details}. As our approach is image-free, training relies exclusively on prompts from JourneyDB \cite{Sun2023JourneyDB}. For both base and distilled models, CFG scales are set to 7.5 (SDXL), 3.5 (PixArt-$\alpha$), 4.5 (SD3.5), and 5.0 (Wan2.1). We employ bilinear interpolation for upsampling. All benchmarks and speed measurements are performed on a single NVIDIA A100 GPU. In our primary experiments, we scale the resolution from $512$px to $1024$px for image synthesis and from 480p ($834 \times 480$) to 720p ($1280 \times 720$) for video generation. This is implemented via a two-stage ($K=2$) multi-resolution strategy, using $\text{logSNR}_1 = -2.5$ as the transition threshold. For instance, in a 4-step SDXL inference, the trajectory is partitioned into a high-noise interval ($t \in [502, 1000]$) processed at low resolution, followed by a refinement stage ($t \in [0, 502]$) at high resolution. Model-specific schedules and further details are provided in Appendix B. The distillation training cost of RMD is comparable to TDM and lower than DMD2 (see Appendix B for detailed comparisons).

\begin{table}[!hbtp]
\centering

\caption{Quantitative comparison of RMD with state-of-the-art methods in text-to-image generation. The best results among distillation methods are highlighted in \textbf{bold}.}
\label{tab:comparison}
\resizebox{\textwidth}{!}{
\begin{tabular}{lcccccccccc} 
\toprule
\multirow{2}{*}{Method} & \multirow{2}{*}{Backbone} & \multirow{2}{*}{NFE} & \multirow{2}{*}{Speedup} & \multicolumn{5}{c}{HPS$\uparrow$} & \multirow{2}{*}{Aes$\uparrow$} & \multirow{2}{*}{CLIP Score$\uparrow$} \\
\cline{5-9}
& & & & Animation & Art & Painting & Photo & Average & & \\

\midrule
\rowcolor{gray!15} Base Model-1024 & SDXL & 40$\times$2 & 1.0$\times$ &27.73 & 31.16& 28.93& 28.73& 29.14& 6.48 & 35.64 \\
SDXL-Turbo-512& SDXL & 4 & 20.0$\times$ & 31.20 & 29.72 & 29.56 & 27.36 & 29.46  & 6.41 & 34.36 \\
SDXL-Lighting  & SDXL & 4 & 20.0$\times$ & 32.46 & 31.07 & 31.23 & 28.57 & 30.83 & 6.54 & 34.62 \\
DMD2 & SDXL & 4 & 20.0$\times$ & 32.75 & 31.79 & 31.79 & 28.87 & 31.30 & 6.71 & 35.10 \\
TDM & SDXL & 4 & 20.0$\times$  & 33.64 & \textbf{32.37}& \textbf{32.18}& 29.51 & 31.93 & 6.70 & 35.03\\
\textbf{RMD (Ours)} & SDXL & $2+2$ & \textbf{33.4$\times$} & \textbf{33.71} & 32.14 & 31.94&  \textbf{30.16} & \textbf{31.99} & \textbf{6.73} & \textbf{35.13} \\
\midrule
\rowcolor{gray!15} Base Model-1024 & PixArt-$\alpha$ & 25$\times2$ & 1.0$\times$ & 32.22& 30.71 & 30.45 & 29.58&30.74 & 6.73 & 34.12\\
YOSO-512  & PixArt-$\alpha$ & 4 & 12.5$\times$ & 31.40 & 31.18 & 31.26 & 28.15 & 30.60 & 6.23 & 31.83 \\
DMD2  & PixArt-$\alpha$ & 4 &12.5$\times$ & 32.56 & 31.57 & 31.57 & 29.76 &31.36  & 6.71 & 33.53 \\
TDM  & PixArt-$\alpha$ & 4 &12.5$\times$ &33.07	& 32.25 & 32.23 & 30.64 & 32.05 & 6.82& 33.63\\
\textbf{RMD (Ours)} & PixArt-$\alpha$ & $2+2$ &\textbf{21.0$\times$} &\textbf{33.24}&	\textbf{32.42}	&\textbf{32.55}	&\textbf{30.71}	& \textbf{32.23} & \textbf{6.89} & \textbf{33.89} \\

\midrule
\rowcolor{gray!15} Base Model-1024& SD3.5& 40$\times$2 & 1.0$\times$& 31.86 & 30.99 & 30.75 & 27.58 & 30.29 & 6.49 & 34.76\\
SD3.5-Turbo & SD3.5 & 4 & 20.0$\times$ &30.07 &28.25 & 28.82& 25.78 &28.23  & 6.35 & \textbf{32.97} \\
DMD2  & SD3.5 & 4 & 20.0$\times$ & 31.38 & 30.47 & 31.20 & 28.65 & 30.42 & 6.32 & 32.16 \\
TDM  & SD3.5 & 4 & 20.0$\times$ & 31.24 & 30.20 & 30.94 & 27.74 & 30.03 & 6.37 & 32.25 \\
\textbf{RMD (Ours)} & SD3.5 & $2+2$ & \textbf{32.0$\times$} & \textbf{31.62}& \textbf{30.50}&\textbf{31.22} &\textbf{28.89} &\textbf{30.56} & \textbf{6.38} & 32.51\\
\bottomrule
\end{tabular}
}

\end{table}

\subsection{Main Results}
\subsubsection{Results on Text-to-Image Generation}
We evaluate the text-to-image generation performance of RMD across multiple representative backbones. We adopt a two-stage inference strategy consisting of two low-resolution steps followed by two high-resolution steps, denoted as an NFE of 2+2. The results are presented in Tab.~\ref{tab:comparison}. From the results, we draw the following observations: (1) Distribution matching methods frequently outperform the base models in certain metrics. This superiority is primarily attributable to the transition from error-prone multi-step trajectories to direct mappings, thereby eliminating cumulative sampling drift. (2) Our method consistently outperforms baselines via cross-resolution distillation. By anchoring core semantic layouts at lower resolutions during early denoising, the model expands its effective receptive field and suppresses high-frequency artifacts. This prioritizes global structural coherence before transitioning to fine-grained local refinement. (3) RMD maintains high fidelity while delivering superior inference efficiency—achieving a 33.4$\times$ acceleration compared to base SDXL and outperforming current step-distillation baselines. Relative to distilled baselines, RMD achieves 1.67$\times$ (SDXL), 1.68$\times$ (PixArt-$\alpha$), 1.60$\times$ (SD3.5), and 1.53$\times$ (Wan2.1) speedup. Qualitative examples are provided in Appendix E.

\begin{table}[!t]
\centering

\caption{Quantitative Results on Text-to-Video Generation. All methods use Wan2.1-T2V-14B as the backbone and are evaluated via VBench and T2V-CompBench (average) using original prompts. The best results among distillation methods are highlighted in \textbf{bold}.}
\label{tab:vbench_summary}
\resizebox{\textwidth}{!}{
\begin{tabular}{lcccccc}
\toprule
Method  & NFE & Speedup & Total score$\uparrow$ & Quality score$\uparrow$  & Semantic score$\uparrow$ & T2V-Compbench$\uparrow$\\
\midrule
\rowcolor{gray!15} Base Model-720p  & 50$\times$2 & 1.0$\times$ & 83.75& 85.44& 76.95  & 54.17\\
DMD2  & 6 & 16.7$\times$ &80.30 & 81.98 & 73.60 & 52.81\\
TDM  & 6 & 16.7$\times$ &80.48&81.85&75.00 & 52.27\\
\textbf{RMD (Ours)}  & $3+3$ & \textbf{25.6$\times$} & \textbf{82.51} & \textbf{84.37}& \textbf{75.05}  & \textbf{54.00}\\
\bottomrule
\end{tabular}
}

\end{table}

\begin{figure}[!t]
  \centering
    \includegraphics[width=1\linewidth]{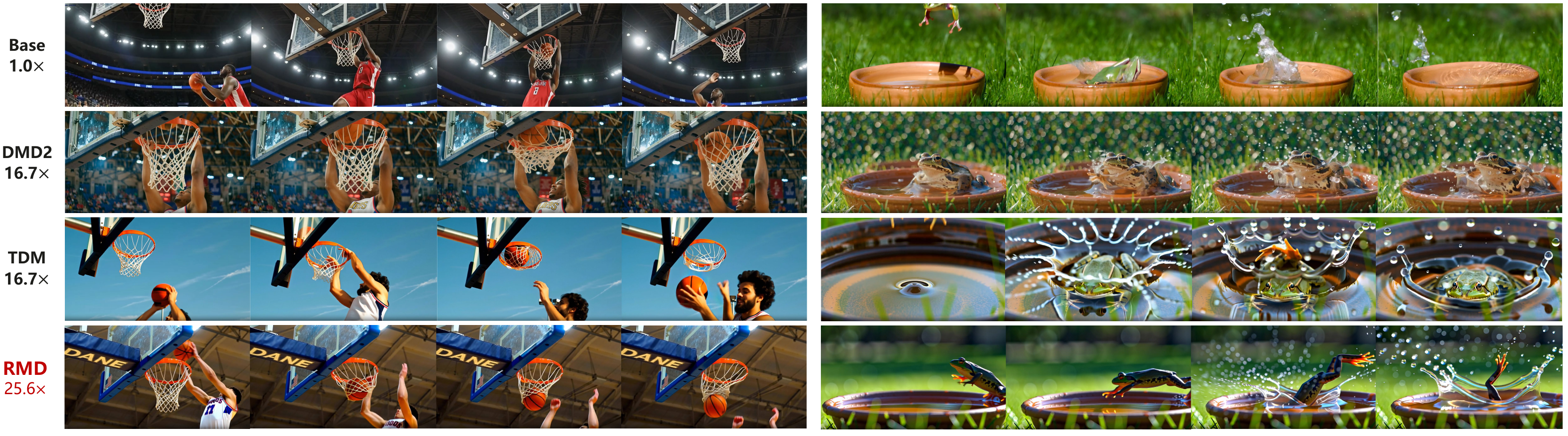}
  \caption{Visual comparison of our model with DMD2, TDM, and the base model.
All distilled models are evaluated with 6 sampling steps, while the teacher model uses 50 steps with classifier-free guidance. For fair comparison, all results are generated using identical noise seeds and text prompts.
Left prompt: A basketball player soaring through the air for a thunderous slam dunk.
Right prompt: A frog jumping into a bowl of water, splashing droplets onto the surrounding grass. }
  \label{fig:video_case_composite} 

\end{figure}

\subsubsection{Results on Text-to-Video.}
We extend RMD to the Wan2.1-T2V-14B as the base model. While SOTA baselines like DMD2 and TDM are constrained to a fixed 6-step NFE, RMD employs a $3+3$ multi-stage strategy. Quantitative results in Tab.~\ref{tab:vbench_summary} show RMD outperforming competitors across all metrics, achieving a 25.6$\times$ speedup—over 50\% more efficient than 6-step baselines. Qualitative comparisons (Fig.~\ref{fig:video_case_composite}) reveal that RMD preserves superior motion details and semantic coherence, whereas DMD2 exhibits limited temporal dynamics. These findings confirm RMD's ability to substantially accelerate video synthesis without compromising quality. 

\begin{table}[!t]
\centering
\caption{Ablation study of the proposed components. Evaluations are conducted on SDXL/Wan2.1 backbone. The best results are highlighted in \textbf{bold}.}
\label{tab:ablation_final}
\begin{tabularx}{\textwidth}{c YY @{\hspace{1em}} c @{\hspace{1em}} YYcY}
\toprule
\multirow{2.5}{*}{Target Resolution} & \multicolumn{2}{c}{Components} & & \multicolumn{4}{c}{Evaluation Metrics} \\
\cmidrule(lr){2-3} \cmidrule(lr){5-8} 
& RM & UP &  & HPS$\uparrow$ & Aes$\uparrow$ & CLIP Score$\uparrow$ & Vbench$\uparrow$ \\
\midrule
\multirow{2}{*}{512px/480p} & $\times$ & $\times$ & & 22.64 & 6.15 & 28.76 & 80.01 \\
& \checkmark & $\times$ & & \textbf{30.28} & \textbf{6.63} & \textbf{34.81} & \textbf{80.81} \\
\midrule
\multirow{2}{*}{1024px/720p} & $\times$ & \checkmark & & 29.57 & 6.58 & 33.68 & 81.11 \\
& \checkmark & \checkmark & & \textbf{31.99} & \textbf{6.73} & \textbf{35.13} & \textbf{82.51}\\
\bottomrule
\end{tabularx}

\end{table}

\begin{figure}[!t]
  \centering
 
  \includegraphics[width=1\linewidth]{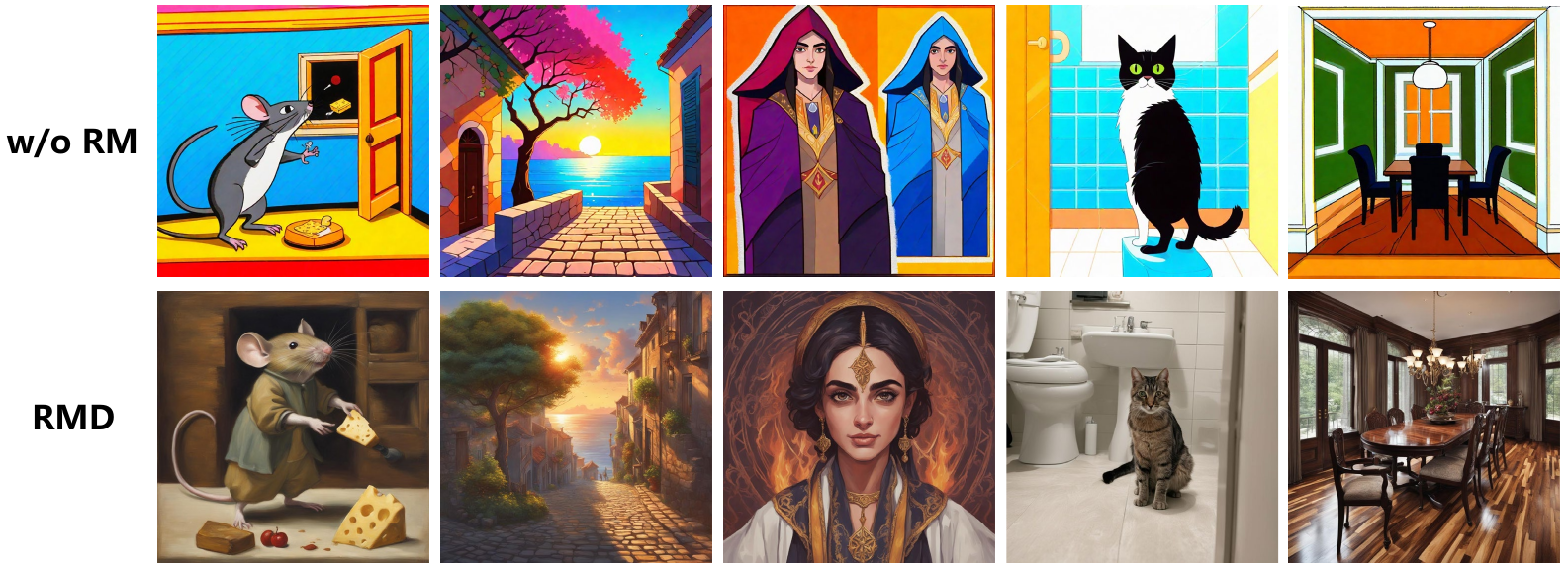} 
  \caption{Visual ablation comparison of models with and without RM on SDXL at a 1024px target resolution. The corresponding prompts are provided in Appendix D.}

  \label{fig:ablation_case}
\end{figure}

\subsection{Ablation Study}
\subsubsection{Component Ablation Analysis.} Tab.~\ref{tab:ablation_final} isolates the effects of the Cross-Resolution Distribution Matching (RM) and Upsampling (UP) modules under a unified evaluation protocol. During training, all models are initialized at a base resolution of 512px/480p, while the target resolution is determined by the UP module, resulting in outputs at 512px/480p or 1024px/720p. Applying RM alone (i.e., without UP) yields substantial improvements over the baseline; these gains primarily arise from RM’s ability to eliminate the distribution gap between low- and high-resolution representations. In contrast, a naive cascaded upsampling strategy without RM performs poorly (see Fig.~\ref{fig:ablation_case}), as the low-resolution stage fails to establish robust semantic structures that can be preserved and refined during subsequent upsampling. When cascaded upsampling is combined with RM, the two components exhibit clear complementarity: UP alleviates the high inference cost at early high-resolution stages, while RM ensures cross-resolution distribution alignment. Together, they achieve the best overall performance.

\subsubsection{Ablation on Noise Re-Injection Strategy.} 
\begin{figure}[!t]
  \centering

  \includegraphics[width=1\linewidth]{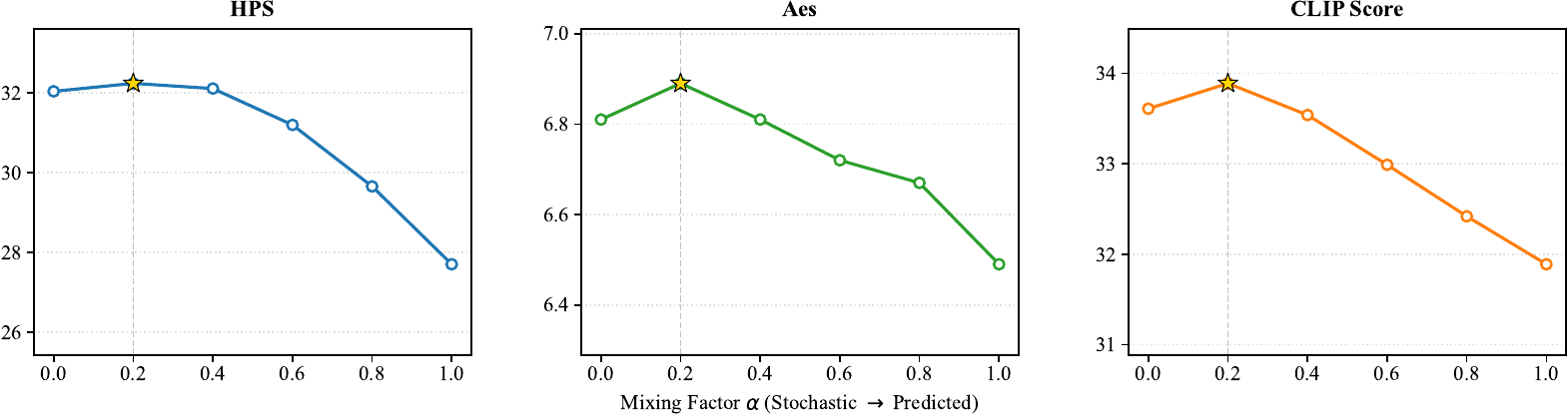} 
  \caption{Ablation study of noise re-injection strategies. Effect of varying the mixing factor $\alpha$, which controls the balance between predicted and stochastic noise. The backbone is PixArt-$\alpha$ (note that $\alpha$ here is part of the model name).}
  \label{fig:ablation_hyper}
  
\end{figure}

We analyze the mixing factor $\alpha$ in Eq.~\eqref{eq:noise_inject} under an NFE = 2+2 multi-resolution schedule (Fig.~\ref{fig:ablation_hyper}). Pure Gaussian noise ($\alpha=0$) enables cross-resolution alignment but fails to inherit the teacher’s ODE trajectory, leading to weak structural consistency. Pure predicted noise ($\alpha=1.0$) strictly follows the teacher trajectory but degrades under large resolution gaps, as upsampling artifacts are treated as hard guidance. The optimal setting $\alpha=0.2$ balances trajectory inheritance and stochastic flexibility, effectively bridging the multi-resolution gap and achieving the best generation fidelity.

\begin{table}[!t]
\centering

\caption{Hyperparameter analysis on $\text{logSNR}$-driven step allocation. The $\text{logSNR}_1$ governs the assignment of sampling steps ($N_{512}$ and $N_{1024}$) across 512px and 1024px resolutions. The backbone is Pixart-$\alpha$.}
\label{tab:hyperparameter_ablation}
\begin{tabularx}{\textwidth}{c c Y Y YYY}
\toprule
\multirow{1}{*}{$N_{512}$ + $N_{1024}$} & \multirow{1}{*}{$\text{logSNR}_1$} & Speedup & HPS$\uparrow$ & Aes$\uparrow$  & CLIP Score$\uparrow$\\
\midrule
0 + 4  & - & 1$\times$ & 32.05 &6.82 & 33.63\\
1 + 3 & -6.00 & 1.25$\times$ & 32.12 & \textbf{6.87} & 33.64 \\
2 + 2 & -2.50 & 1.68$\times$ & \textbf{32.13} & 6.83 & \textbf{33.86}\\
3 + 1 & 0.00 & \textbf{2.54$\times$} & 24.79 & 6.03 & 32.22\\

\bottomrule
\end{tabularx}
\end{table}

\subsection{Analysis of Resolution Trajectory Division}
\begin{figure}[!t]
  \centering
  \includegraphics[width=1\linewidth]{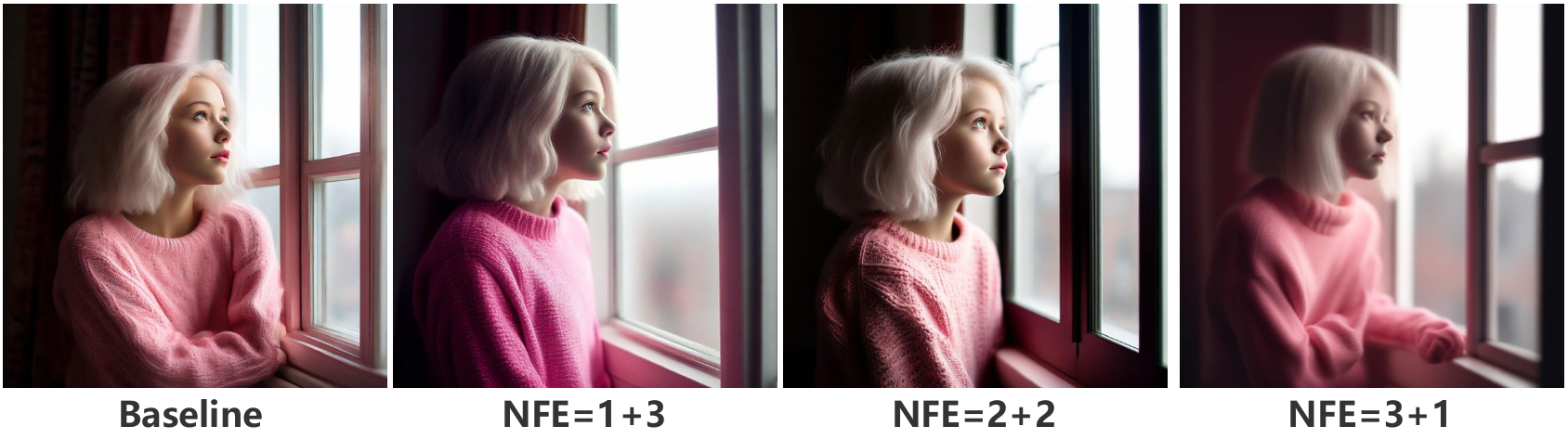} 
  \caption{Qualitative comparison of different step allocations. Prompt: A white-haired girl in a pink sweater looks out a window in her bedroom.}
  \label{fig:hyperparameter_analysis}
\end{figure}

We analyze the impact of the threshold $\text{logSNR}_1$ on the distribution of sampling steps across scales. As shown in Tab.~\ref{tab:hyperparameter_ablation}, decreasing $\text{logSNR}_1$ shifts the computational budget toward the lower resolution, yielding a higher speedup by mitigating the quadratic attention complexity at 1024px. The configuration with $N_{512}=2$ and $N_{1024}=2$ achieves the best balance between efficiency and quality, delivering a 1.68$\times$ speedup while attaining the highest HPS (32.13) and CLIP (33.86) scores. While the $1+3$ configuration offers a minor improvement in aesthetic score (6.87), this marginal gain is achieved at the expense of a significant reduction in inference speed. Conversely, compressing the high-resolution refinement to a single step ($3+1$) leads to a sharp degradation in quality. This observation is visually substantiated in Fig.~\ref{fig:hyperparameter_analysis}, where the $2+2$ assignment preserves intricate details and structural sharpness, whereas the $3+1$ scheme manifests as significant blurriness and a total loss of fine textures. Consequently, we adopt $\text{logSNR}_1=-2.5$ as the default setting for our model to achieve an optimal balance between efficiency and generative fidelity.
\begin{table}[!t]
\centering
\caption{Effect of multi-stage resolution training with RMD on SDXL. We evaluate progressive resolution schedules (256px $\rightarrow$ 512px $\rightarrow$ 1024px), corresponding to $\text{logSNR}_1$ and $\text{logSNR}_2$, which induce different NFE allocations across stages. The backbone model is SDXL. $^\ddagger$ NFE per stage; omitted if 1.}

\label{tab:multi_stage_results}
\setlength{\tabcolsep}{5pt} 

\begin{tabularx}{\textwidth}{l >{\small}c >{\small}c c Y Y Y}
\toprule
Method & Resolution (NFE)$^\ddagger$ & $\text{logSNR}_1, \text{logSNR}_2$ & Speedup & HPS$\uparrow$ & Aes$\uparrow$ & CLIP$\uparrow$ \\ \midrule
\multirow{2}{*}{Baseline} & {\shortstack[c]{1024(4)}} & -- & 1.00$\times$ & 31.93 & 6.70 & 35.03 \\ 
 & {\shortstack[c]{512(2)$\to$1024(2)}} & -2.5 & 1.56$\times$ & 29.57 & 6.58 & 33.68 \\ \cmidrule{1-7}
\multirow{5}{*}{RMD} & {\shortstack[c]{512(2)$\to$1024(2)}} & -2.5 & 1.56$\times$ & \textbf{31.99} & \textbf{6.73} & \textbf{35.13} \\ \cmidrule{2-7}
 & {\shortstack[c]{256$\to$512$\to$1024(2)}} & {\shortstack{-6.0,-2.5}} & 1.73$\times$ & 31.59 & 6.70 & 34.84 \\
 & {\shortstack[c]{256$\to$512(2)$\to$1024}} & \shortstack{-6.0,-0.0} & 2.54$\times$ & 31.48 & 6.68 & 34.77 \\
 & {\shortstack[c]{256(2)$\to$512$\to$1024}} & \shortstack{-2.5,-0.0} & 2.90$\times$ & 31.03 & 6.67 & 34.57 \\ \cmidrule{2-7}
 & {\shortstack[c]{128$\to$256$\to$512$\to$1024}} & \shortstack{-6.0, -2.5, -0.0} & \textbf{3.08$\times$} & 29.48 & 6.59 & 33.41 \\ 
\bottomrule
\end{tabularx}

\end{table}

\subsection{Effectiveness of Multi-stage Resolution Training} 
The evaluation of our resolution progressive strategy on SDXL, summarized in Tab.~\ref{tab:multi_stage_results}, demonstrates that RMD effectively optimizes the trade-off between generative quality and inference speed. Specifically, the 2-stage RMD configuration ($2+2$ NFE) outperforms the single-resolution baseline across all metrics—achieving an HPS of 31.99 and an Aesthetic score of 6.73—while providing a 1.56$\times$ speedup. This performance significantly exceeds a naive 2-stage baseline (HPS 29.57), confirming that our RM module successfully bridges the distribution gap during resolution transitions. Furthermore, extending this to a 3-stage 256px $\rightarrow$ 512px $\rightarrow$ 1024px strategy enables even greater efficiency, reaching a 2.90$\times$ speedup with the $2+1+1$ NFE allocation while maintaining favorable visual fidelity. These results validate the scalability of our coarse-to-fine distillation logic, proving it can achieve high-quality synthesis with substantially reduced computational overhead.

\section{Conclusions}
We propose a cross-resolution distribution matching distillation (RMD) framework for few-step, coarse-to-fine multi-resolution cascaded generation. RMD explicitly resolves distribution mismatch across resolutions and enables progressive resolution expansion within a limited number of inference steps, overcoming the efficiency bottleneck of conventional step-reduction distillation while preserving high-fidelity synthesis. Extensive experiments on image and video generation validate its effectiveness. RMD consistently achieves state-of-the-art sampling speed without sacrificing visual quality, providing a scalable solution for accelerating future generative models.

\section{Acknowledgements}
This work was supported by TokensEngine 1.0 under Grant No. 9473277, the National Natural Science Foundation of China under Grant No. 62276131, the Natural Science Foundation of Jiangsu Province under Grant No. BK20240081, and the Special Research Project on Teaching Reform of General Artificial Intelligence Courses in Jiangsu Undergraduate Universities under Grant No. ZNT-10.
\bibliographystyle{splncs04}
\bibliography{main}

@inproceedings{Ho2020DDPM,
  title={Denoising Diffusion Probabilistic Models},
  author={Ho, Jonathan and Jain, Ajay and Abbeel, Pieter},
  booktitle={NeurIPS},
  year={2020}
}

@inproceedings{Chen2023NoiseSchedule,
  title={On the Importance of Noise Scheduling for Diffusion Models},
  author={Chen, Ting},
  booktitle={ICML},
  year={2023}
}

@inproceedings{Hoogeboom2023SimpleDiffusion,
  title={Simple Diffusion: End-to-End Diffusion for High Resolution Images},
  author={Hoogeboom, Emiel and Heek, Jonathan and Salimans, Tim},
  booktitle={ICLR},
  year={2023}
}

@article{Luo2025TDM,
  title={Learning Few-Step Diffusion Models by Trajectory Distribution Matching},
  author={Luo, Yihong and Hu, Tianyang and Sun, Jiacheng and Cai, Yujun and Tang, Jing},
  journal={arXiv preprint arXiv:2503.06674},
  year={2025}
}

@article{Tian2025BottleneckSampling,
  title={Training-free Diffusion Acceleration with Bottleneck Sampling},
  author={Tian, Ye and Xia, Xin and Ren, Yuxi and Lin, Shanchuan and Wang, Xing and Xiao, Xuefeng and Tong, Yunhai and Yang, Ling and Cui, Bin},
  journal={arXiv preprint arXiv:2503.18940},
  year={2025}
}

@article{Esser2024SD3,
  title={Scaling Rectified Flow Transformers for High-Resolution Image Synthesis},
  author={Esser, Patrick and Kulal, Sumith and Blattmann, Andreas and Entezari, Rahim and M{\"u}ller, Jonas and others},
  journal={arXiv preprint arXiv:2403.03206},
  year={2024}
}

@article{Chen2025PixelFlow,
  title={PixelFlow: Pixel-Space Generative Models with Flow},
  author={Chen, Shoufa and Ge, Chongjian and Zhang, Shilong and Sun, Peize and Luo, Ping},
  journal={arXiv preprint arXiv:2504.07963},
  year={2025}
}

@article{Ma2025NAMI,
  title={NAMI: Efficient Image Generation via Bridged Progressive Rectified Flow Transformers},
  author={Ma, Yuhang and Cheng, Bo and Liu, Shanyuan and Zhou, Hongyi and Wu, Xiaoyu and Wu, Liebucha and Leng, Dawei and Yin, Yuhui},
  journal={arXiv preprint arXiv:2503.09242},
  year={2025}
}

@article{Hunyuan2025Video,
  title={HunyuanVideo: A Systematic Framework for Large Video Generative Models},
  author={Tencent Hunyuan Team},
  journal={arXiv preprint},
  year={2025}
}

@article{Wan2025Video,
  title={Wan: Open and Advanced Large-Scale Video Generative Models},
  author={Wan Team},
  journal={arXiv preprint},
  year={2025}
}

@inproceedings{Sauer2024ADD,
  title        = {Adversarial Diffusion Distillation},
  author       = {Axel Sauer and Dominik Lorenz and Andreas Blattmann and Robin Rombach},
  booktitle    = {ECCV},
  year         = {2024},
}

@misc{Shan2024PADD,
  title        = {SDXL-Lightning: Progressive Adversarial Diffusion Distillation},
  author       = {Shanchuan Lin and Anran Wang and Xiao Yang},
  howpublished = {CoRR},
  year         = {2024},
}

@inproceedings{PodellELBDMPR2024SDXL,
  title        = {{SDXL:} Improving Latent Diffusion Models for High-Resolution Image Synthesis},
  author       = {Dustin Podell and Zion English and Kyle Lacey and Andreas Blattmann and Tim Dockhorn and Jonas M{\"{u}}ller and Joe Penna and Robin Rombach},
  booktitle    = {ICLR},
  year         = {2024},
}

@inproceedings{Chen2024Pixart,
  title        = {PixArt-{\(\alpha\)}: Fast Training of Diffusion Transformer for Photorealistic
                  Text-to-Image Synthesis},
  author       = {Junsong Chen and Jincheng Yu and Chongjian Ge and Lewei Yao and Enze Xie and Zhongdao Wang and James T. Kwok and Ping Luo and Huchuan Lu and Zhenguo Li},
  booktitle    = {ICLR},
  year         = {2024},
}

@inproceedings{LuoCQHT25YOSO,
  title        = {You Only Sample Once: Taming One-Step Text-to-Image Synthesis by Self-Cooperative Diffusion GANs},
  author       = {Yihong Luo and Xiaolong Chen and Xinghua Qu and Tianyang Hu and Jing Tang},
  booktitle    = {ICLR},
  year         = {2025}, 
}

@misc{Wu2023HPS,
  title        = {Human Preference Score v2: {A} Solid Benchmark for Evaluating Human Preferences of Text-to-Image Synthesis},
  author       = {Xiaoshi Wu and Yiming Hao and Keqiang Sun and Yixiong Chen and Feng Zhu and Rui Zhao and Hongsheng Li},
  howpublished = {CoRR},
  year         = {2023},
}

@misc{Schuhmann2022Aes,
      title={LAION-5B: An open large-scale dataset for training next generation image-text models}, 
      author={Christoph Schuhmann and Romain Beaumont and Richard Vencu and Cade Gordon and Ross Wightman and Mehdi Cherti and Theo Coombes and Aarush Katta and Clayton Mullis and Mitchell Wortsman and Patrick Schramowski and Srivatsa Kundurthy and Katherine Crowson and Ludwig Schmidt and Robert Kaczmarczyk and Jenia Jitsev},
      journal={arXiv preprint},
      year={2022},
}

@inproceedings{Huang2024Vbench,
  title        = {VBench: Comprehensive Benchmark Suite for Video Generative Models},
  author       = {Ziqi Huang and Yinan He and Jiashuo Yu and Fan Zhang and Chenyang Si and Yuming Jiang and Yuanhan Zhang and Tianxing Wu and Qingyang Jin and Nattapol Chanpaisit and Yaohui Wang and Xinyuan Chen and Limin Wang and Dahua Lin and Yu Qiao and Ziwei Liu},
  booktitle    = {CVPR},
  year         = {2024},
}

@inproceedings{Sun2025Compbench,
  author       = {Kaiyue Sun and Kaiyi Huang and Xian Liu and Yue Wu and Zihan Xu and Zhenguo Li and Xihui Liu},
  title        = {T2V-CompBench: {A} Comprehensive Benchmark for Compositional Text-to-video Generation},
  booktitle    = {CVPR},
  year         = {2025},
}

@inproceedings{Sun2023JourneyDB,
  title        = {JourneyDB: {A} Benchmark for Generative Image Understanding},
  author       = {Keqiang Sun and Junting Pan and Yuying Ge and Hao Li and Haodong Duan and Xiaoshi Wu and Renrui Zhang and Aojun Zhou and Zipeng Qin and Yi Wang and Jifeng Dai and Yu Qiao and Limin Wang and Hongsheng Li},
  booktitle    = {NeurIPS},
  year         = {2023},
}

@inproceedings{Rombach2022LDM,
  author       = {Robin Rombach and
                  Andreas Blattmann and
                  Dominik Lorenz and
                  Patrick Esser and
                  Bj{\"{o}}rn Ommer},
  title        = {High-Resolution Image Synthesis with Latent Diffusion Models},
  booktitle    = {CVPR},
  pages        = {10674--10685},
  year         = {2022},
}

@inproceedings{Peebles2023DiT,
  author       = {William Peebles and
                  Saining Xie},
  title        = {Scalable Diffusion Models with Transformers},
  booktitle    = {ICCV},
  pages        = {4172--4182},
  year         = {2023},
}

@inproceedings{Liu2025TE-Cache,
  author       = {Feng Liu and
                  Shiwei Zhang and
                  Xiaofeng Wang and
                  Yujie Wei and
                  Haonan Qiu and
                  Yuzhong Zhao and
                  Yingya Zhang and
                  Qixiang Ye and
                  Fang Wan},
  title        = {Timestep Embedding Tells: It's Time to Cache for Video Diffusion Model},
  booktitle    = {CVPR},
  pages        = {7353--7363},
  year         = {2025},
}

@inproceedings{Song2021DDIM,
  author       = {Jiaming Song and
                  Chenlin Meng and
                  Stefano Ermon},
  title        = {Denoising Diffusion Implicit Models},
  booktitle    = {ICLR},
  year         = {2021},
}

@article{Lu2025DPM-Solver++,
  author       = {Cheng Lu and
                  Yuhao Zhou and
                  Fan Bao and
                  Jianfei Chen and
                  Chongxuan Li and
                  Jun Zhu},
  title        = {DPM-Solver++: Fast Solver for Guided Sampling of Diffusion Probabilistic
                  Models},
  journal      = {Mach. Intell. Res.},
  year         = {2025},
}

@inproceedings{Song2023CM,
  author       = {Yang Song and
                  Prafulla Dhariwal and
                  Mark Chen and
                  Ilya Sutskever},
  title        = {Consistency Models},
  booktitle    = {ICML},
  year         = {2023},
}

@inproceedings{Yin2024DMD2,
  author       = {Tianwei Yin and
                  Micha{\"{e}}l Gharbi and
                  Taesung Park and
                  Richard Zhang and
                  Eli Shechtman and
                  Fr{\'{e}}do Durand and
                  Bill Freeman},
  title        = {Improved Distribution Matching Distillation for Fast Image Synthesis},
  booktitle    = {NeurIPS},
  year         = {2024},
}

@inproceedings{Yin2024DMD,
  author       = {Tianwei Yin and
                  Micha{\"{e}}l Gharbi and
                  Richard Zhang and
                  Eli Shechtman and
                  Fr{\'{e}}do Durand and
                  William T. Freeman and
                  Taesung Park},
  title        = {One-Step Diffusion with Distribution Matching Distillation},
  booktitle    = {CVPR},
  year         = {2024},
}

@inproceedings{Zhou2024SiD,
  author       = {Mingyuan Zhou and
                  Huangjie Zheng and
                  Zhendong Wang and
                  Mingzhang Yin and
                  Hai Huang},
  title        = {Score identity Distillation: Exponentially Fast Distillation of Pretrained
                  Diffusion Models for One-Step Generation},
  booktitle    = {ICML},
  volume       = {235},
  pages        = {62307--62331},
  year         = {2024},
}

@inproceedings{Salimans2022PD,
  author       = {Tim Salimans and
                  Jonathan Ho},
  title        = {Progressive Distillation for Fast Sampling of Diffusion Models},
  booktitle    = {ICLR},
  year         = {2022},
}

@inproceedings{Kim2024CTM,
  author       = {Dongjun Kim and
                  Chieh{-}Hsin Lai and
                  Wei{-}Hsiang Liao and
                  Naoki Murata and
                  Yuhta Takida and
                  Toshimitsu Uesaka and
                  Yutong He and
                  Yuki Mitsufuji and
                  Stefano Ermon},
  title        = {Consistency Trajectory Models: Learning Probability Flow {ODE} Trajectory
                  of Diffusion},
  booktitle    = {ICLR},
  year         = {2024},
}

@misc{Wang2023Painter,
  author       = {Binxu Wang and
                  John J. Vastola},
  title        = {Diffusion Models Generate Images Like Painters: an Analytical Theory
                  of Outline First, Details Later},
  howpublished = {CoRR, abs/2303.02490},
  year         = {2023},
}

@inproceedings{Yuan2024DiTFastAttn,
  author       = {Zhihang Yuan and
                  Hanling Zhang and
                  Lu Pu and
                  Xuefei Ning and
                  Linfeng Zhang and
                  Tianchen Zhao and
                  Shengen Yan and
                  Guohao Dai and
                  Yu Wang},
  title        = {DiTFastAttn: Attention Compression for Diffusion Transformer Models},
  booktitle    = {NeurIPS},
  year         = {2024},
}

@inproceedings{Lipman2023FM,
  author       = {Yaron Lipman and
                  Ricky T. Q. Chen and
                  Heli Ben{-}Hamu and
                  Maximilian Nickel and
                  Matthew Le},
  title        = {Flow Matching for Generative Modeling},
  booktitle    = {ICLR},
  year         = {2023},
}

@article{Shen2025Survey,
  author       = {Hui Shen and
                  Jingxuan Zhang and
                  Boning Xiong and
                  Rui Hu and
                  Shoufa Chen and
                  Zhongwei Wan and
                  Xin Wang and
                  Yu Zhang and
                  Zixuan Gong and
                  Guangyin Bao and
                  Chaofan Tao and
                  Yongfeng Huang and
                  Ye Yuan and
                  Mi Zhang},
  title        = {Efficient Diffusion Models: {A} Survey},
  journal      = {Trans. Mach. Learn. Res.},
  volume       = {2025},
  year         = {2025},
}

@inproceedings{Liu2024InstaFlow,
  author       = {Xingchao Liu and
                  Xiwen Zhang and
                  Jianzhu Ma and
                  Jian Peng and
                  Qiang Liu},
  title        = {InstaFlow: One Step is Enough for High-Quality Diffusion-Based Text-to-Image
                  Generation},
  booktitle    = {ICLR},
  year         = {2024},
}

@inproceedings{Wang2024PCM,
  author       = {Fu{-}Yun Wang and
                  Zhaoyang Huang and
                  Alexander William Bergman and
                  Dazhong Shen and
                  Peng Gao and
                  Michael Lingelbach and
                  Keqiang Sun and
                  Weikang Bian and
                  Guanglu Song and
                  Yu Liu and
                  Xiaogang Wang and
                  Hongsheng Li},
  title        = {Phased Consistency Models},
  booktitle    = {NeurIPS},
  year         = {2024},
}

@misc{Jiang2025RLDMD,
  author       = {Dengyang Jiang and
                  Dongyang Liu and
                  Zanyi Wang and
                  Qilong Wu and
                  Liuzhuozheng Li and
                  Hengzhuang Li and
                  Xin Jin and
                  David Liu and
                  Zhen Li and
                  Bo Zhang and
                  Mengmeng Wang and
                  Steven C. H. Hoi and
                  Peng Gao and
                  Harry Yang},
  title        = {Distribution Matching Distillation Meets Reinforcement Learning},
  howpublished = {CoRR, abs/2511.13649},
  year         = {2025},
}

@misc{Xu2025fDMD,
  author       = {Yilun Xu and
                  Weili Nie and
                  Arash Vahdat},
  title        = {One-step Diffusion Models with \emph{f}-Divergence Distribution Matching},
  howpublished = {CoRR, abs/2502.15681},
  year         = {2025},
}

@misc{Bandyopadhyay2025SD35Flash,
  author       = {Hmrishav Bandyopadhyay and
                  Rahim Entezari and
                  Jim Scott and
                  Reshinth Adithyan and
                  Yi{-}Zhe Song and
                  Varun Jampani},
  title        = {SD3.5-Flash: Distribution-Guided Distillation of Generative Flows},
  howpublished = {CoRR},
  year         = {2025},
}

@misc{Lu2025ADMD,
  author       = {Yanzuo Lu and
                  Yuxi Ren and
                  Xin Xia and
                  Shanchuan Lin and
                  Xing Wang and
                  Xuefeng Xiao and
                  Andy J. Ma and
                  Xiaohua Xie and
                  Jian{-}Huang Lai},
  title        = {Adversarial Distribution Matching for Diffusion Distillation Towards
                  Efficient Image and Video Synthesis},
  howpublished = {CoRR, abs/2507.18569},
  year         = {2025},
}

@inproceedings{Meng2023GuidedDistill,
  author       = {Chenlin Meng and
                  Robin Rombach and
                  Ruiqi Gao and
                  Diederik P. Kingma and
                  Stefano Ermon and
                  Jonathan Ho and
                  Tim Salimans},
  title        = {On Distillation of Guided Diffusion Models},
  booktitle    = {CVPR},
  pages        = {14297--14306},
  year         = {2023},
}

@inproceedings{Song2024ITC,
  author       = {Yang Song and
                  Prafulla Dhariwal},
  title        = {Improved Techniques for Training Consistency Models},
  booktitle    = {ICLR},
  year         = {2024},
}

@misc{wu2026diversity,
  title={Diversity-Preserved Distribution Matching Distillation for Fast Visual Synthesis},
  author={Wu, Tianhe and Li, Ruibin and Zhang, Lei and Ma, Kede},
  howpublished = {CoRR, abs/2602.03139},
  year={2026}
}

@article{goodfellow2020generative,
  title={Generative adversarial networks},
  author={Goodfellow, Ian and Pouget-Abadie, Jean and Mirza, Mehdi and Xu, Bing and Warde-Farley, David and Ozair, Sherjil and Courville, Aaron and Bengio, Yoshua},
  journal={Communications of the ACM},
  volume={63},
  number={11},
  pages={139--144},
  year={2020},
}

@misc{Zhang2025Waver,
  author       = {Yifu Zhang and
                  Hao Yang and
                  Yuqi Zhang and
                  Yifei Hu and
                  Fengda Zhu and
                  Chuang Lin and
                  Xiaofeng Mei and
                  Yi Jiang and
                  Bingyue Peng and
                  Zehuan Yuan},
  title        = {Waver: Wave Your Way to Lifelike Video Generation},
  howpublished = {CoRR, abs/2508.15761},
  year         = {2025},
}

@misc{Cai2025LongCat,
  author       = {Xunliang Cai and
                  Qilong Huang and
                  Zhuoliang Kang and
                  Hongyu Li and
                  Shijun Liang and
                  Liya Ma and
                  Siyu Ren and
                  Xiaoming Wei and
                  Rixu Xie and
                  Tong Zhang},
  title        = {LongCat-Video Technical Report},
  howpublished = {CoRR},
  year         = {2025},
}

@inproceedings{Teng2024Relay,
  author       = {Jiayan Teng and
                  Wendi Zheng and
                  Ming Ding and
                  Wenyi Hong and
                  Jianqiao Wangni and
                  Zhuoyi Yang and
                  Jie Tang},
  title        = {Relay Diffusion: Unifying diffusion process across resolutions for
                  image synthesis},
  booktitle    = {ICLR},
  year         = {2024},
}

@inproceedings{Liu2023FSF,
  author       = {Xingchao Liu and
                  Chengyue Huang and
                  Xiaojie Lin and
                  Yang Wang},
  title        = {Flow Straight and Fast: Learning to Generate and Transfer Data with Rectified Flow},
  booktitle    = {ICLR},
  year         = {2023},
}

@misc{Zhang2025DTC,
  author       = {Yuechen Zhang and
                  Jinbo Xing and
                  Bin Xia and
                  Shaoteng Liu and
                  Bohao Peng and
                  Xin Tao and
                  Pengfei Wan and
                  Eric Lo and
                  Jiaya Jia},
  title        = {Training-Free Efficient Video Generation via Dynamic Token Carving},
  howpublished = {CoRR, abs/2505.16864},
  year         = {2025},
}

@misc{Zhang2025FlashVideo,
  author       = {Shilong Zhang and
                  Wenbo Li and
                  Shoufa Chen and
                  Chongjian Ge and
                  Peize Sun and
                  Yida Zhang and
                  Yi Jiang and
                  Zehuan Yuan and
                  Binyue Peng and
                  Ping Luo},
  title        = {FlashVideo: Flowing Fidelity to Detail for Efficient High-Resolution
                  Video Generation},
  howpublished = {CoRR, abs/2502.05179},
  year         = {2025},
}

@inproceedings{SDEdit,
  author       = {Chenlin Meng and
                  Yutong He and
                  Yang Song and
                  Jiaming Song and
                  Jiajun Wu and
                  Jun{-}Yan Zhu and
                  Stefano Ermon},
  title        = {SDEdit: Guided Image Synthesis and Editing with Stochastic Differential
                  Equations},
  booktitle    = {ICLR},
  year         = {2022},
}

@inproceedings{Diff-Instruct,
  author       = {Weijian Luo and
                  Tianyang Hu and
                  Shifeng Zhang and
                  Jiacheng Sun and
                  Zhenguo Li and
                  Zhihua Zhang},
  editor       = {Alice Oh and
                  Tristan Naumann and
                  Amir Globerson and
                  Kate Saenko and
                  Moritz Hardt and
                  Sergey Levine},
  title        = {Diff-Instruct: {A} Universal Approach for Transferring Knowledge From
                  Pre-trained Diffusion Models},
  booktitle    = {NeurIPS},
  year         = {2023},
}

\clearpage
\setcounter{section}{0}
\renewcommand{\thesection}{\Alph{section}}
\begin{center}
{\LARGE\bfseries Appendix}
\end{center}
\section{Distillation and inference Algorithm}

Algorithm 1 presents the pseudocode of Cross-Resolution Distribution Matching Distillation (RMD). It illustrates how a pretrained diffusion model with $T$-step inference is distilled into a fast generator that performs $N$-step inference through a $K$-stage multi-resolution cascade. The procedure details the trajectory partitioning, cross-resolution alignment, and stage-wise optimization strategy that enable efficient compression while preserving distributional consistency across resolutions.
\begin{algorithm}
\caption{RMD: Cross-Resolution Distribution Matching Distillation}
\label{alg:rmd_eccv}
\begin{algorithmic}[1]

\Require Teacher $s_\varphi$, Generator $G_\theta$, Fake score $s_\phi$
\Require Resolutions $\{r_i\}_{i=1}^{K}$, Timesteps $\{t_j\}_{j=0}^{N-1}$
\Require Weights $\{\lambda_{r_i}\}$, Learning rate $\ell$
\Ensure Optimized generator $G_\theta$ and fake score $s_\phi$

\State Initialize $G_\theta$ and $s_\phi$ from teacher $s_\varphi$
\State Divide diffusion trajectory into $K$ resolution-specific intervals $[T_{i-1}, T_i]$

\While{training}
    \State \textbf{// Generate multi-resolution distribution}
    \State Sample base noise $\epsilon \sim \mathcal{N}(0, I)$ at resolution $r_1$
    \State $\{t_{i,j}\}_{j=0}^{N-1} \gets Shift(\{r_{i}\}_{i=1}^{K},  \{t_{j}\}_{j=0}^{N-1})$ \Comment{resolution-timestep mapping}
    \State Generate $\{\tilde{x}_{t_{i,j}}\}_{j=0}^{N-1}$ from $\epsilon$ via cascaded sampling using $G_\theta$

    \State \textbf{// Sample distribution according to the resolution}
    \State $B \gets \lfloor K/2 \rfloor \text{ if warm-up phase, else } K$ \Comment{Warm-up stages}

    \State Sample resolution $\{r_i\}_{i=1}^{B}$ and corresponding timesteps $\{t_i\}_{i=1}^{B}$
    \State Extract states $\{\tilde{x}_{t_i}\}_{i=1}^{B}$ from multi-resolution distribution
    \State Upsample to final resolution: $x_t^{(t_i)} \gets \tilde{x}_{t_i}$ \Comment{Eq.~(10)}

    \State \textbf{// Update fake score}
    \State $L_\phi \gets \lambda_{\text{snr}} \| s_\phi(x_t^{(t_i)}, t) - U_{r_K}(\tilde{x}_0^{(t_i)}) \|_2^2$ \Comment{Eq.~(9)}
    \State $s_\phi \gets s_\phi - \ell \nabla_\phi L_\phi$

    \State \textbf{// Update generator}
    \State $L_\theta \gets \lambda_{r_i} \| x_t^{(t_i)} - \operatorname{sg}(x_t^{(t_i)} + s_\phi - s_\varphi) \|_2^2$ \Comment{Eq.~(8)}
    \State $G_\theta \gets G_\theta - \ell \nabla_\theta L_\theta$

\EndWhile
\end{algorithmic}
\end{algorithm}

Algorithm 2 presents the pseudocode for the $K$-stage multi-resolution cascaded inference with $N$ sampling steps. It describes how the inference process dynamically determines whether to perform upsampling based on the subsequent timestep, and how noise is re-injected after upsampling to maintain consistency with the diffusion trajectory. This design enables seamless resolution transitions while preserving the generative dynamics across stages.
\begin{algorithm}[]
\caption{Multi-Resolution Cascaded Inference}
\label{alg:mr_cascade_inference}
\begin{algorithmic}[1]

\State \textbf{Input:} Generator $G_\theta$, Resolutions $\{r_i\}_{i=1}^{K}$
\State \textbf{Input:} Timesteps $\{t_j\}_{j=0}^{N-1}$ with $t_0=T_0$, $t_{N-1}=T_K$
\State \textbf{Output:} Final sample $x_0$ at resolution $r_K$

\State Initialize noise $\epsilon \sim \mathcal{N}(0,I)$ at resolution $r_1$

\For{$j = 0$ to $N-1$}
    \State Map timestep to resolution: $\tau_j \gets Shift(r_i, t_j)$
    \Comment{Simply $t_{i,j}$ as $\tau_j$}
    \State Predict noise: $\epsilon_\theta \gets \epsilon_\theta(\tilde{x}_{\tau_j}, \tau_j)$

    \If{$\tau_{j+1} \in [T^{(r_i)}_{i-1}, T^{(r_i)}_{i}]$}
        \State Denoise:
        $\tilde{x}_{\tau_{j+1}} \gets \mathrm{Denoise}(\tilde{x}_{\tau_j}, \epsilon_\theta, \tau_j)$
        \Comment{Within same resolution}
    \Else
        \State Recover clean latent:
        $\tilde{x}_0^{(\tau_j)} \gets \tilde{x}_{\tau_j} - \sigma_{\tau_j}\epsilon_\theta$
        \State Upsample and re-inject noise:
        \[
        \tilde{x}_{\tau_{j+1}} \gets
        (1-\sigma_{\tau_{j+1}}) U_{r_{i+1}}(\tilde{x}_0^{(\tau_j)})
        + \sigma_{\tau_{j+1}} \epsilon
        \]
        \State Increment resolution: $i \gets i + 1$
        \Comment{Move to next resolution stage}
    \EndIf
\EndFor

\State \Return $x_0 \gets \tilde{x}_{t_N}$

\end{algorithmic}
\end{algorithm}
\vspace{-10pt}
\section{Experiment details}
\label{sec:implental_details}
\noindent \textbf{Pseudo-Huber Metric.}
Following prior findings (e.g., iCT) that the Pseudo-Huber metric yields superior stability and robustness over the $\ell_2$ metric, we optimize Eq. (8) with the Pseudo-Huber objective. Consistent with standard practice for $d$-dimensional data, the hyperparameter is set to $C = 0.00054\sqrt{d}$.

\noindent \textbf{Implementation Details.} For SDXL, Pixart-$\alpha$, and SD3.5, we employ the AdamW optimizer ($\beta_1=0, \beta_2=0.999$) with a batch size of 32. The generator and fake score utilize constant learning rates of $4 \times 10^{-6}$ and $2 \times 10^{-5}$, respectively. For Wan2.1, we adopt a modified configuration with $\beta_2=0.95$, a batch size of 4, and learning rates of $5 \times 10^{-6}$ ($LR_g$) and $2 \times 10^{-5}$ ($LR_{fake}$). In all experiments, gradient norm clipping is applied with a threshold of 1.0. To facilitate a stable initialization, a warmup strategy is implemented during low-resolution distillation, consisting of 50 steps for SD-based models and 20 steps for Wan2.1. Regarding the noise prediction intensity $\alpha$, we set it to 0.2 and 0.5 for the respective models during training to modulate the learning objectives. During inference, $\alpha$ is adjusted to 1.0 for the SD series and 0.9 for Wan2.1.

\noindent \textbf{Timestep and Resolution Schedules.}
In our primary experiments, we define specific schedules for the timestep and resolution.
Following the Flow Shift strategy, we apply a shift scale of $3$ for SD3.5 and $5$ for Wan2.1.
The detailed mapping between the shifted timesteps and their corresponding resolution stages is summarized in Tab.~\ref{tab:schedules}.
\setcounter{table}{5}
\begin{table}[!t]
\centering
\caption{Timestep and resolution schedules. \textbf{Bold} values indicate shifted timesteps adjusted via the logSNR timestep mapping. Original timesteps are shown in parentheses following each shifted value.}
\label{tab:schedules}
\resizebox{\columnwidth}{!}{
    \begin{tabularx}{1.2\columnwidth}{@{}l l l@{}} 
    \toprule
    Model  & Timestep Schedule & Resolution \\ \midrule
    SDXL/PixArt & [1000, \textbf{681} (750), 500, 250] & [512px, 512px, 1024px, 1024px] \\ \addlinespace[2pt]
    SD3.5 & [1000, \textbf{865} (900), 750, 500] & [512px, 512px, 1024px, 1024px]\\ \addlinespace[2pt]
    Wan2.1-14B & [1000, \textbf{937} (962), \textbf{857} (909), 834, 716, 505] & [480p, 480p, 480p, 720p, 720p, 720p] \\ \bottomrule
    \end{tabularx}
}
\end{table}
\setcounter{figure}{8}
\begin{figure}[!t]
\centering
\includegraphics[width=0.8\linewidth]{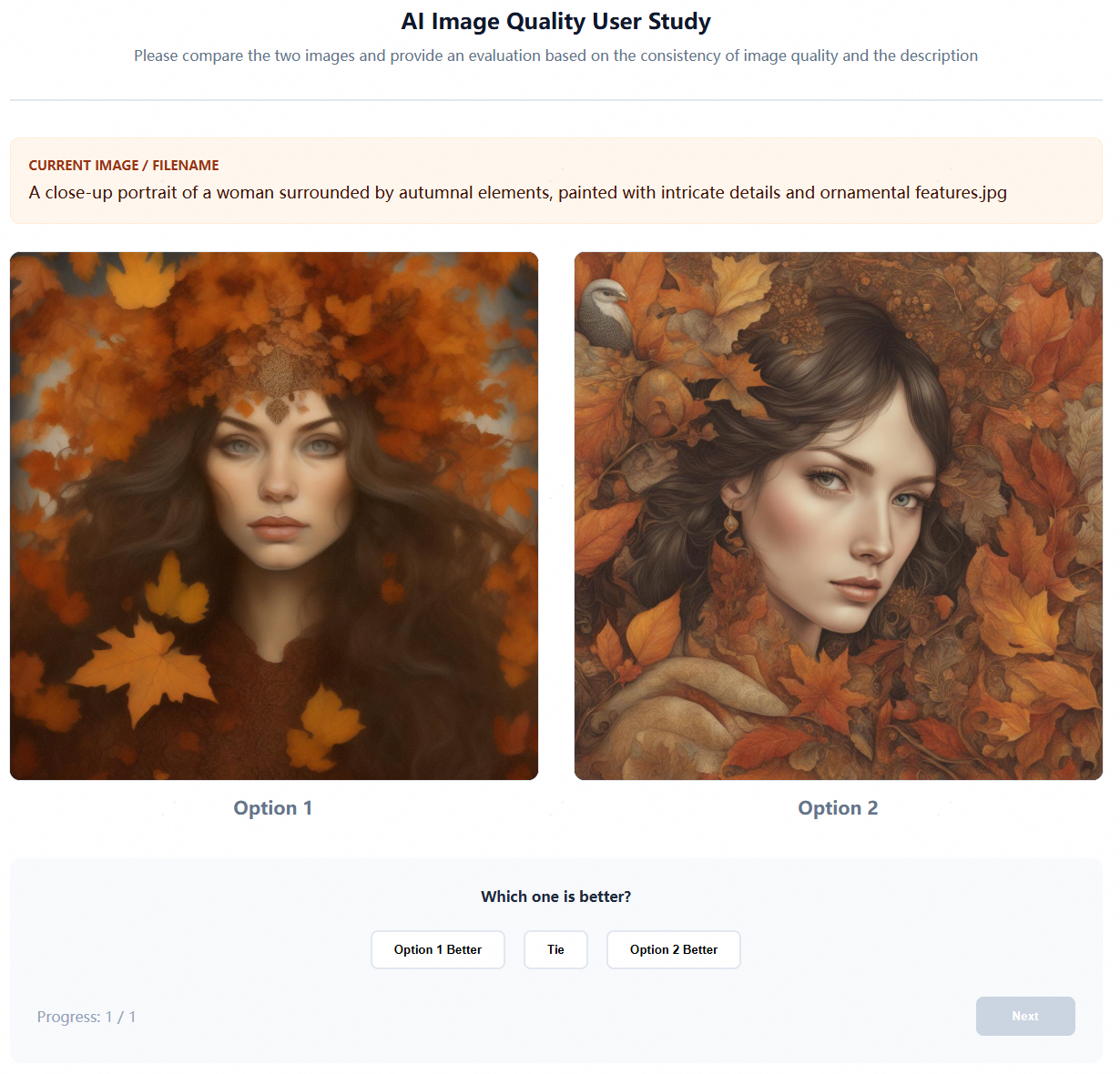} 
\caption{Example of the human evaluation interface used in our user study.}
\label{fig:demo}
\end{figure}

\begin{figure}[!t]
\centering
\includegraphics[width=0.8\linewidth]{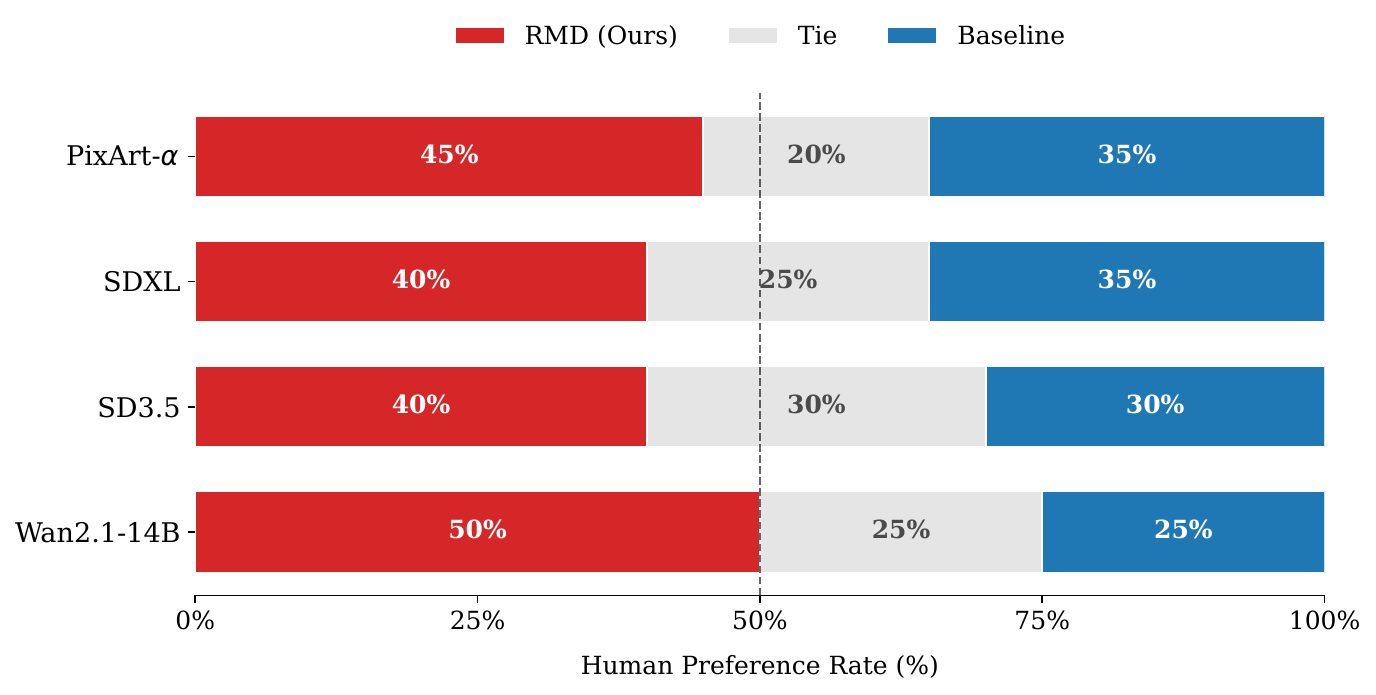}
\caption{\textbf{User preference results} comparing RMD and Baseline (TDM) across four base models. RMD shows a consistent advantage in human-perceived image quality and prompt alignment.}
\label{fig:user_study}
\end{figure}

\section{User Study}
We conducted a human evaluation to compare the generation quality of RMD and Baseline (TDM).
Participants were presented with pairs of anonymous images generated by both models and asked to select the one with superior visual fidelity and text-to-image alignment.
For each of the four base models (PixArt-$\alpha$, SDXL, SD3.5, and Wan2.1-14B), we randomly selected 20 prompts for evaluation. All generated samples were manually screened to exclude inappropriate or sensitive content.
An example of the evaluation interface is illustrated in Fig.~\ref{fig:demo}.
In total, we collected and analyzed responses from multiple independent evaluators, providing a representative insight into human preference across various architectures.
The results, summarized in Fig.~\ref{fig:user_study}, indicate that RMD consistently achieves a higher preference rate than TDM across all tested models, albeit by a slight margin, demonstrating its superior capability in preserving fine-grained details.
%
%
\section{Prompts}
\label{appendix:prompts}

The textual prompts for the visual results in Fig. 1 and Fig. 6 are listed below, following the left-to-right sequence in each figure.

\subsection*{Prompts for Figure 1}
\begin{enumerate}[label=(\arabic*), leftmargin=2em, nosep]
    \item A realistic painting of a bifurcated astronaut suit with a clear brain case and camera appendage stalks, covered in diamond and iridescent fractal bubble materials, in a jumping float pose.
    \item A lemon character with sunglasses on the beach.
    \item A pirate with a beer is illustrated in detailed digital painting.
    \item Lionel Messi portrayed as a sitcom character.
    \item A monk in an orange robe looks out of a round window in a spaceship in dramatic lighting.
\end{enumerate}

\subsection*{Prompts for Figure 6}
\begin{enumerate}[label=(\arabic*), leftmargin=2em, nosep]
    \item Museum painting of a mouse stealing cheese artwork.
    \item A cobblestone street with a tree over the sea at sunset, illuminated by sun rays, in a colorful illustration by Peter Chan on Artstation.
    \item Maya Ali as a D\&D Mage wearing wizard robes in the style of various artists, depicted in a head-on symmetrical painted portrait.
    \item A cat standing on a toilet seat in a bathroom.
    \item A dining room with hard wood floors that is very fancy.
\end{enumerate}

\section{Visual Results}
\begin{figure}[!t]
    \centering
    \includegraphics[width=\linewidth]{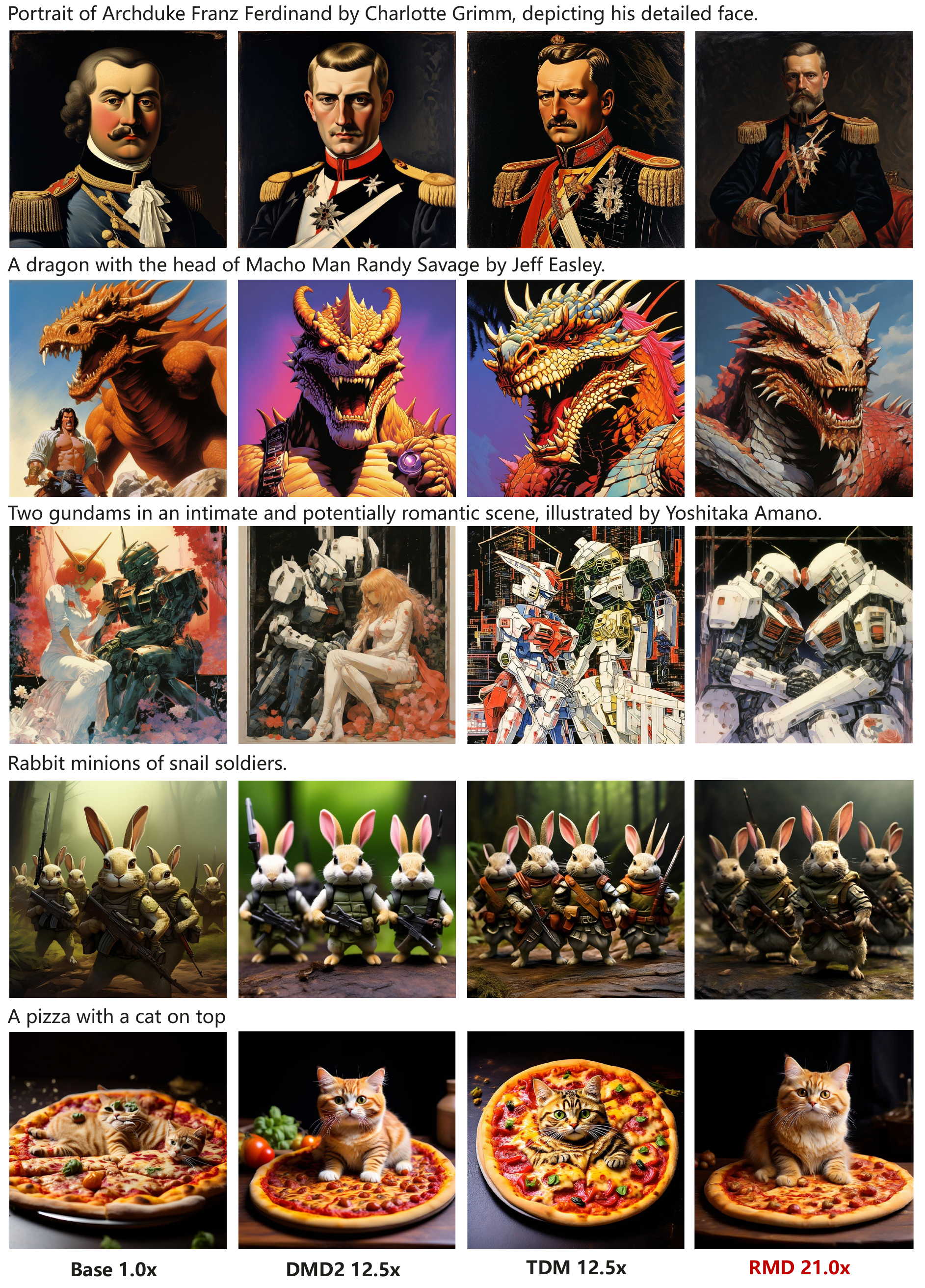}
    \caption{Visual comparison on PixArt-$\alpha$ using identical seeds and prompts.}
    \label{fig:app_pixart}
\end{figure}

\begin{figure}[!t]
    \centering
    \includegraphics[width=\linewidth]{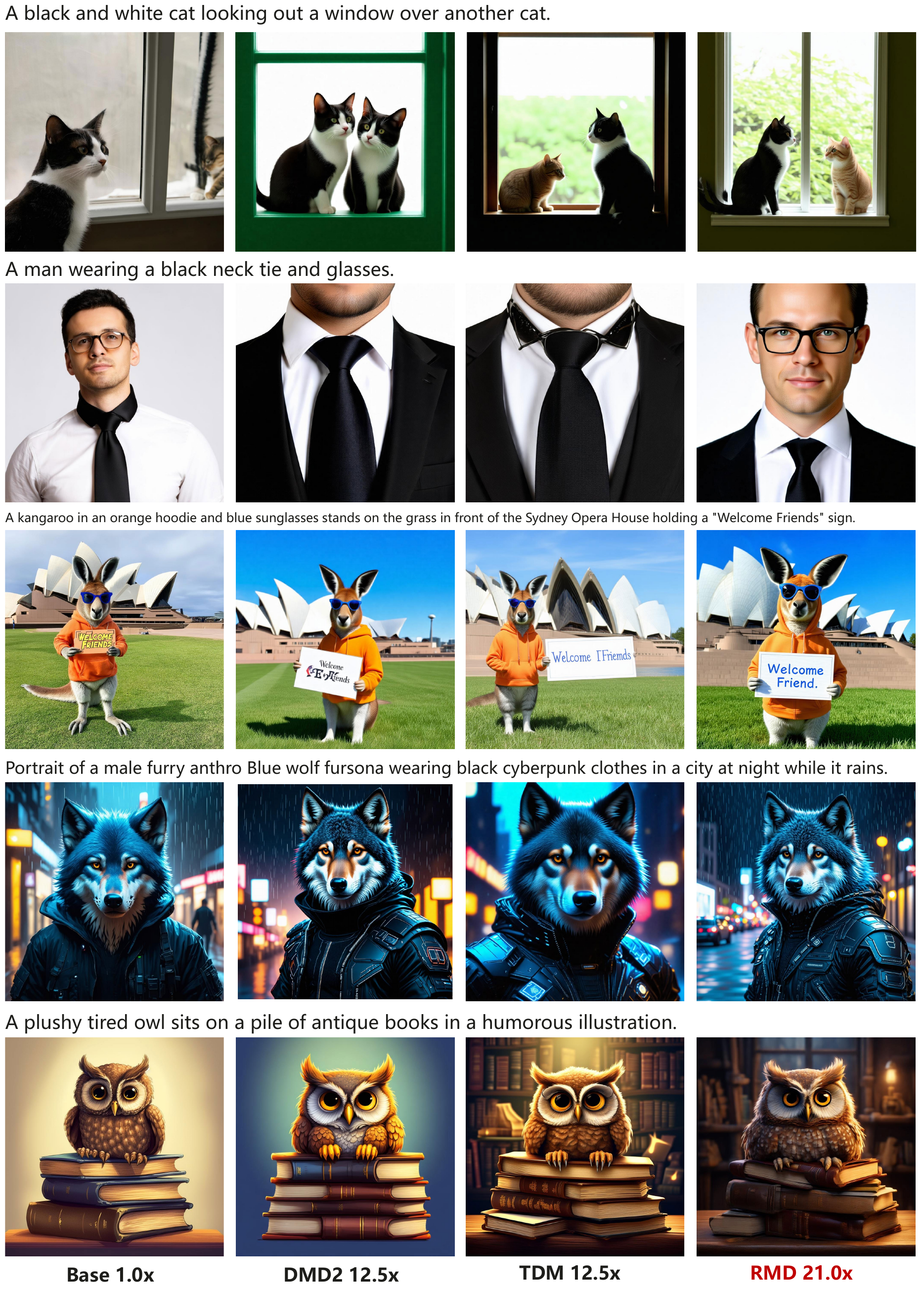}
    \caption{Visual comparison on SD3.5 using identical seeds and prompts.}
    \label{fig:app_sd35}
\end{figure}

\begin{figure}[!t]
      \centering
      \begin{subfigure}[t]{\linewidth}
          \centering
          \includegraphics[width=\linewidth]{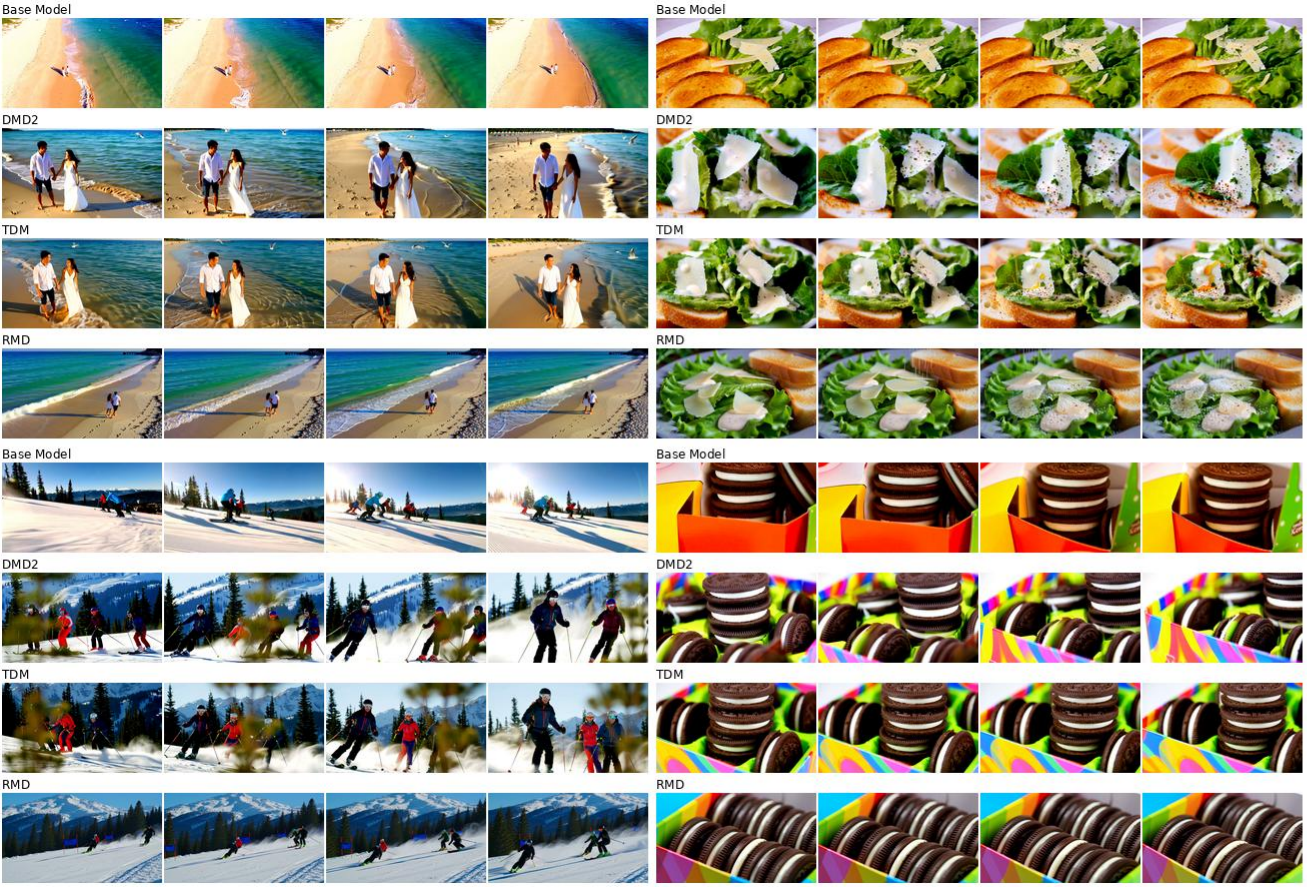}
          \caption{}
          \label{fig:app_video_1}
      \end{subfigure}
      \\[\abovecaptionskip]
      \begin{subfigure}[t]{\linewidth}
          \centering
          \includegraphics[width=\linewidth]{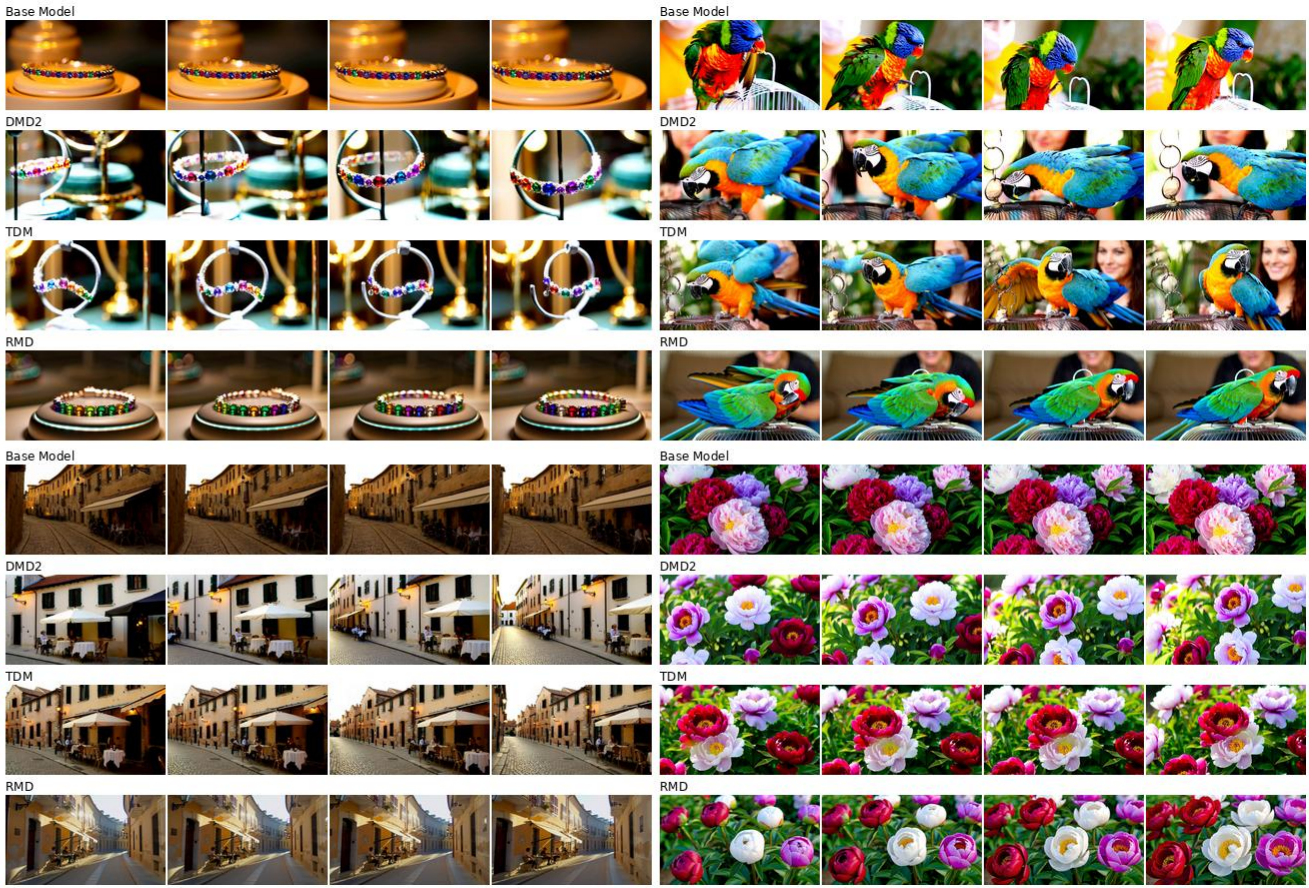}
          \caption{}
          \label{fig:app_video_2}
      \end{subfigure}
      \caption{Additional video comparison results on Wan2.1-14B.}
      \label{fig:app_video}
  \end{figure}

Additional visual comparisons in Figs.~\ref{fig:app_pixart} and~\ref{fig:app_sd35} show that our RMD maintains high-fidelity synthesis across both SD3.5 and PixArt-$\alpha$ backbones, achieving significant acceleration with only 4 sampling steps. Further video comparisons on Wan2.1-14B are shown in Fig.~\ref{fig:app_video}.

\end{document}